\documentclass[11pt, a4paper, logo, copyright, nonumbering]{cls/Sci}

\usepackage[numbers, sort&compress, square]{natbib}
\usepackage{url}
\usepackage{amsmath}
\usepackage[table]{xcolor}
\usepackage{multirow}
\usepackage{makecell}
\usepackage{booktabs}
\usepackage{amsfonts}
\usepackage{nicefrac}
\usepackage{graphicx}
\usepackage{enumitem}
\usepackage{pifont}
\usepackage{pdflscape}
\usepackage{array}
\usepackage{caption}
\usepackage{subcaption}
\usepackage{wrapfig}
\usepackage{float}
\usepackage{placeins}
\usepackage{listings}
\usepackage{adjustbox}
\usepackage{fontawesome5}
\usepackage{textcomp}
\usepackage{etoolbox}
\usepackage[most]{tcolorbox}

\newcommand{\cmark}{\textcolor{green}{\ding{51}}}
\newcommand{\xmark}{\textcolor{red}{\ding{55}}}

\lstdefinestyle{python}{
  language=Python,
  basicstyle=\ttfamily\small,
  keywordstyle=\color{blue}\bfseries,
  stringstyle=\color{red!70!black},
  commentstyle=\color{green!50!black}\itshape,
  showstringspaces=false,
  breaklines=true,
  frame=single,
  framerule=0.4pt,
  rulecolor=\color{black!30},
  backgroundcolor=\color{black!3},
  numbers=left,
  numberstyle=\tiny\color{black!40},
  numbersep=6pt,
  tabsize=4,
  literate=
    {->}{{\textrightarrow}}2
    {None}{{\textbf{None}}}1
    {True}{{\textbf{True}}}1
    {False}{{\textbf{False}}}1,
}
\title{DataMaster: Data-Centric Autonomous AI Research}

\author{
{\normalfont\mdseries
Yaxin Du\textsuperscript{1,*}, 
Xiyuan Yang\textsuperscript{1,*}, 
Zhifan Zhou\textsuperscript{2}, 
Wanxu Liu\textsuperscript{1}, 
Zixing Lei\textsuperscript{1}, 
Zimeng Chen\textsuperscript{1}, 
Fenyi Liu\textsuperscript{1},
Haotian Wu\textsuperscript{1}, 
Yuzhu Cai\textsuperscript{4}, 
Zexi Liu\textsuperscript{1}, 
Xinyu Zhu\textsuperscript{1}, 
Wenhao Wang\textsuperscript{3}, 
Linfeng Zhang\textsuperscript{1}, 
Chen Qian\textsuperscript{1}, 
Siheng Chen\textsuperscript{1,\ensuremath{\dagger}}
}
\\
{\normalfont\bfseries
\textsuperscript{1}Shanghai Jiao Tong University \quad
\textsuperscript{2}Carnegie Mellon University \quad
\textsuperscript{3}Zhejiang University
}
\\
{\normalfont\bfseries
\textsuperscript{4}Beijing University of Aeronautics and Astronautics
}
\\
{\normalfont\mdseries
\textsuperscript{*}Equal contribution.
\quad
\textsuperscript{\ensuremath{\dagger}}Corresponding author: sihengc@sjtu.edu.cn
}

}

\begin{document}

\begin{abstract}
As model families, training recipes, and compute budgets become increasingly standardized, further gains in machine learning systems depend increasingly on data. Yet data engineering remains largely manual and ad hoc: practitioners repeatedly search for external datasets, adapt them to existing pipelines, validate candidate data through downstream training, and carry forward lessons from prior attempts.
We study \textbf{task-conditioned autonomous data engineering}, where an autonomous agent improves a fixed learning algorithm by optimizing only the data side, including external data discovery, data selection and composition, cleaning and transformation. The goal is to obtain a stronger downstream solution while leaving the learning algorithm unchanged.
To address the open-ended search space, branch-dependent refinement, and delayed validation inherent in autonomous data engineering, we propose \textbf{DataMaster}, a data-agent framework that integrates tree-structured search, shared candidate data, and cumulative memory. DataMaster consists of three key components: a \emph{DataTree} that organizes alternative data-engineering branches, a shared \emph{Data Pool} that stores discovered external data sources for reuse, and a \emph{Global Memory} that records node outcomes, artifacts, and reusable findings. Together, these components allow the agent to discover candidate data, construct executable training inputs, evaluate them through downstream feedback, and carry useful evidence across branches. We evaluate DataMaster on two types of benchmarks, MLE-Bench Lite and PostTrainBench. 
On MLE-Bench Lite, it improves medal rate by 32.27\% over the initial score; on PostTrainBench, it surpasses the instruct model on GPQA (31.02\% vs 30.35\%).

{\normalfont\mdseries
\rule{0pt}{1.55em}\faGithub\quad
\textbf{Code}\quad
\url{https://github.com/sjtu-sai-agents/DataMaster}
}

\end{abstract}

\maketitle

\section{Introduction}
\label{sec:intro}

Progress in Machine Learning (ML) has historically been driven by better models, better algorithms, and more computation.
Yet in an increasing number of settings, model development itself is becoming more standardized: pretrained backbones are widely available, training recipes are easier to reuse, open-source stacks make fine-tuning increasingly turnkey, and evaluation pipelines are more reproducible than before.
As a result, performance differences are increasingly shaped by data: what data is used, how it is selected and cleaned, how it is mixed, and how well it matches the downstream objective.
This points to a more strongly data-centric phase of ML development~\citep{zha2023datacentricartificialintelligencesurvey,mazumder2023dataperfbenchmarksdatacentricai}. It also changes what autonomous systems should optimize: if data is becoming the main remaining lever for improving downstream performance, then agentic ML systems should move beyond model-centric optimization alone.

In practice, current autonomous ML systems remain largely model-centric. Recent ML engineering agents have become increasingly capable at writing code, debugging training pipelines, tuning hyperparameters, and iterating from validation feedback, achieving strong results in benchmarked environments such as MLE-Bench~\citep{chan2025mlebenchevaluatingmachinelearning,jiang2025aide,zhu2026ultralonghorizonagenticsciencecognitive}. More recent benchmarks such as PostTrainBench have begun to expand the action space by allowing agents to go online, run experiments, and curate data during training~\citep{rank2026posttrainbenchllmagentsautomate}. However, even in these settings, data is still treated primarily as an auxiliary action dimension rather than the central object of optimization. What remains missing is a framework that allows an agent to systematically improve downstream performance by operating on the data side itself.
To address this gap, we study \textit{task-conditioned autonomous data engineering} for ML systems.
Given a downstream objective and an initial solution, the agent treats the \emph{data state}, rather than the model code, as the primary object of optimization and iteratively searches, cleans, adapts, integrates, and reuses data under training-closed-loop feedback.
The objective is not to perform an isolated data operation, but to improve the full data pipeline, including discovering candidate sources, cleaning and adapting them, integrating them into training, and validating whether they improve the downstream model. 
Existing data-centric work addresses important parts of this pipeline, either through autonomous systems for tasks such as dataset discovery~\citep{li2025datasetresearchbenchmarkingagentsystems} and data cleaning~\citep{gao2025closingdataloopusing}, or through benchmarks that evaluate these stages in isolation~\citep{cai2025opendataarenafairopenarena,lai2026kramabenchbenchmarkaisystems}. However, these stages remain fragmented and are rarely linked by downstream training feedback.
This problem is harder than simply retrieving a few extra datasets and appending them to the training set.
\textbf{\ding{182} Data search is open-ended}: potentially useful sources are scattered across repositories and the open web, with uneven quality, unclear licensing, and highly variable relevance to the task.
\textbf{\ding{183} Discovered data is rarely plug-and-play}: schema alignment, label mapping, format conversion, filtering, and deduplication often have to be synthesized before the data can even enter training.
\textbf{\ding{184} Data iteration is inherently cumulative}.
To avoid repeating failed directions and to build on past successes, the agent must remember where it searched, what transformations were attempted, which adapters worked, and what kinds of data improved performance.
To instantiate this idea, we introduce \textbf{DataMaster}, the first data-agent framework that treats the \emph{data state} as the primary optimization target and unifies external data discovery, executable data refinement, and cumulative memory under downstream training feedback. Its core search structure, \textbf{DataTree}, organizes data engineering as tree-structured search over self-contained data states, each of which includes the current training data and its associated transformations, paired with a shared \textbf{Data Pool} that stores discovered external datasets for reuse and a \textbf{Global Memory} that stores reusable source histories, useful cleaning methods, and training outcomes across nodes. This tree-plus-memory design is natural for data engineering because data iteration is branchable, path-dependent, cumulative, and only verifiable through downstream training. Concretely, DataTree decomposes data engineering into two complementary node types: \emph{red nodes}, which search for and introduce external data, and \emph{black nodes}, which clean, adapt, and enhance the current training data. Together, DataTree with Global Memory turn autonomous data engineering from a sequence of isolated data operations into a structured search process over executable data states, enabling the agent to systematically improve downstream performance through data-side decisions.
We evaluate DataMaster in two complementary settings: MLE-Bench Lite for controlled classical ML engineering and PostTrainBench for autonomous LLM post-training. On MLE-Bench Lite, DataMaster improves the strongest initial solution by +32.27\% in medal rate. On PostTrainBench, DataMaster improves the average score by +19.79 points over the strongest non-human baseline and surpasses official instruct model on GPQA (31.02 vs. 30.35). These results show that autonomous data engineering can deliver substantial gains across both traditional ML engineering and LLM post-training.

The main contributions of this paper are as follows:

\begin{itemize}[leftmargin=*,topsep=0pt,itemsep=3pt,parsep=0pt,partopsep=0pt]
    \item We formulate \emph{task-conditioned autonomous data engineering} as a first-class research problem, in which the data state rather than model code is the primary object of optimization.
    \item We introduce \textbf{DataMaster}, a memory-augmented data-agent framework whose \textbf{DataTree} organizes autonomous data engineering as tree-structured search over data states, paired with a \textbf{Global Memory} for cumulative reuse across nodes.
    \item We evaluate DataMaster on two complementary benchmarks, \textbf{PostTrainBench} and \textbf{MLE-Bench Lite}, showing that structured data-side iteration can improve strong initial solutions across both LLM post-training and classical ML engineering settings.
\end{itemize}




\section{Related Work}
\label{sec:related}

\textbf{Data agents and data-centric AI}
The data-centric AI paradigm~\citep{zha2023datacentricartificialintelligencesurvey} motivates a line of work that builds autonomous agents to handle data engineering tasks.
CAAFE~\citep{hollmann2023caafe} uses LLMs to generate semantically meaningful features for tabular datasets, focusing on feature-level transformation of a \emph{fixed} input table.
Co-STEER~\citep{yang2024collaborative} extends this to a full data-centric development loop, focusing on evolving the scheduling and implementation of data operations over a \emph{given} dataset.
Dataforge~\citep{wang2025dataforge} focuses on cleaning and optimizing tabular data through dual feedback loops; LAMBDA~\citep{Sun_2025_lambda} focuses on iterative code generation and debugging for data analysis; and DatawiseAgent~\citep{zhang2025datawiseagent} focuses on orchestrating the full notebook-centric data science workflow end-to-end.
Benchmarks such as Spider2-V~\citep{cao2024spider2v} and KramaBench~\citep{lai2026kramabenchbenchmarkaisystems} reveal that these agents still struggle with complex, heterogeneous real-world pipelines.
Across these works, the data sources are \emph{given in advance}, the agent's job is to process and transform on the given dataset.
DataMaster differs in that the source space itself is open-ended: the agent must discover, evaluate, and adapt external data from scratch, and use downstream training performance as the sole quality signal to guide this process.

\noindent \textbf{ML engineering agents and automated learning}
A growing body of work builds LLM-based agents that autonomously run ML experiments.
AIDE~\citep{jiang2025aide} and SELA~\citep{lin2024sela} frame ML engineering as a tree search over code solutions, focusing on optimizing model architecture, hyperparameters, and training scripts.
MLZero~\citep{feng2025mlzero}, MLE-STAR~\citep{nam2025mlestar}, and ML-Master~\citep{zhu2026ultralonghorizonagenticsciencecognitive} further push this direction with multi-agent memory, web-search-guided model retrieval, and hierarchical cognitive caching, all targeting stronger performance on MLE-Bench~\citep{chan2025mlebenchevaluatingmachinelearning}.
Across all of these systems, training data is treated as a \emph{fixed given}---the agent's action space is confined to model code, architecture, and hyperparameters, and data itself is never optimized.
PostTrainBench~\citep{rank2026posttrainbenchllmagentsautomate} takes a step toward data-awareness by allowing agents to curate data during post-training, but data curation remains an auxiliary action rather than the central optimization target. Recent progress tends to focus on this data topic~\citep{huggingface2026mlintern}. 
DataMaster is designed for the complementary regime: the model code is fixed, and the agent's entire action space operates on the \emph{data state}, discovering external sources, synthesizing data processing pipelines, and accumulating memory across iterations, with downstream training performance as the sole optimization signal.

\section{Method}
\label{sec:method}

\subsection{Problem Formulation}

We introduce a new problem: \emph{task-conditioned autonomous data engineering}. Given a downstream ML task and a fixed learning algorithm, an autonomous agent aims to maximize downstream task performance by operating only on the data side. Specifically, the agent may perform \emph{data-side operations}, including external dataset discovery, data selection and composition, cleaning and transformation, and feature construction, while the algorithm itself remains unchanged.

To make this problem precise, we define a \emph{data state} $D$ as the executable configuration produced by such data-side operations under the fixed learning algorithm. It specifies the available data sources, their selected and composed training subsets, and the processing logic needed to convert them into the final inputs consumed by the algorithm.
The input to the problem consists of a downstream task $T$, a fixed learning algorithm $a_0$, an initial executable data state $D_0$, and a resource budget $B$ on downstream training and evaluation, measured in time or compute. For any candidate data state $D$, we use $E(T,a_0,D)$ to denote the downstream performance. The agent searches over the set $\Pi_B(D_0)$ of executable data states reachable from initial data state $D_0$ under budget $B$, and terminates at a refined data state $D^*$ with the best downstream performance. Formally, 
{\setlength{\abovedisplayskip}{3pt}
\setlength{\belowdisplayskip}{2pt}
\begin{equation}
D^* = \arg\max_{D \in \Pi_B(D_0)} E(T, a_0, D),
\label{eq:objective}
\end{equation}}
where the budget $B$ is computing source or time constraints and $D^*$ is used as the final data configuration consumed by the fixed algorithm $a_0$ to produce the downstream model for task $T$.

This problem is challenging for two fundamental reasons. First, it is \emph{open-world}: useful data may come from arbitrary external repositories and the web, so the search space is not explicitly enumerable in advance. Second, it is \emph{long-horizon}: the value of a data-side decision often becomes clear only after subsequent refinement and actual downstream training. As a result, solving this problem requires both broad exploration over candidate data sources and cumulative memory over prior discoveries, transformations, and outcomes.

\begin{figure}
    \centering
    \includegraphics[width=\linewidth]{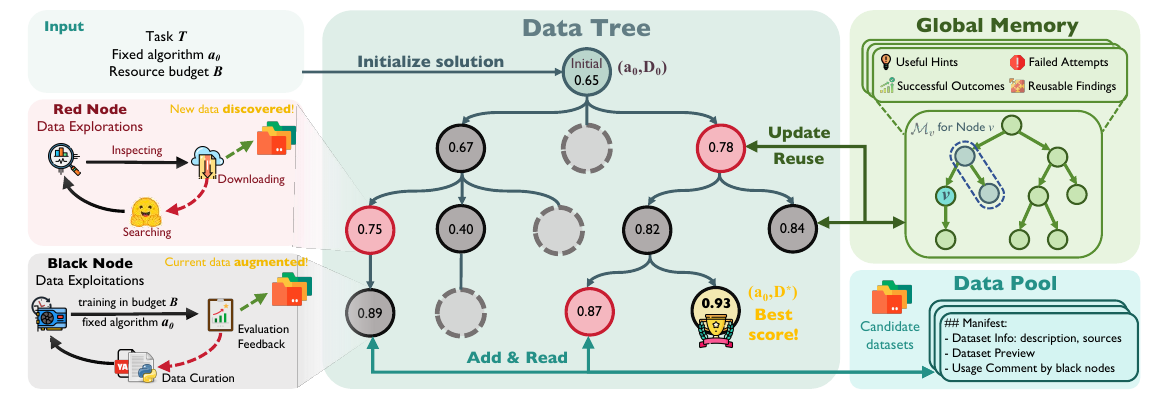}
\caption{
Overview of \textbf{DataMaster}. DataMaster organizes autonomous data engineering as a \textbf{DataTree}, where \emph{red nodes} broaden the search by discovering external datasets and writing them into a shared \textbf{Data Pool}, while \emph{black nodes} exploit available candidates to construct executable data states and obtain downstream training feedback. \textbf{Global Memory} stores reusable artifacts, node outcomes, and prior findings, enabling later nodes to reuse discovered data, avoid repeated failures, and coordinate search across branches under a limited budget.
}
\label{fig:main}
\end{figure}
\subsection{Framework Overview}
\label{subsec:framework_overview}

The formulation above requires searching over a large and open-ended space of reachable executable data states. DataMaster approximates this search by letting an autonomous agent iteratively construct, evaluate, and improve data states for the fixed downstream algorithm. Instead of performing one-shot dataset curation, the agent incrementally searches for new data, tests different refinement choices, and updates later decisions using downstream training feedback. This requires three capabilities: trying multiple next-step choices from the same current state, sharing discovered external datasets across branches, and reusing what earlier attempts have already learned.

DataMaster realizes this through three coupled components: a DataTree, a shared Data Pool, and a Global Memory, shown in Fig.~\ref{fig:main}. The \textbf{DataTree} is the core tree-structured search component of DataMaster, organizing alternative data-engineering branches while preserving branch-local context for later data-side decisions.
The \textbf{Data Pool} is a global set of candidate external datasets discovered so far; red nodes write newly discovered entries into it, and black nodes read from it during refinement. The \textbf{Global Memory} stores reusable processing histories, materialized artifacts, training outcomes, and node-level findings so that later nodes can build on earlier discoveries rather than restarting from scratch.
In addition, due to the limited budget, the tree \textbf{growth and scheduling mechanisms} make the search efficient: downstream training feedback guides how the tree is expanded, while the scheduling policy decides which frontier nodes should be executed first.
Together, these components and their collaboration enable broad data exploration without restarting each branch from scratch, allowing DataMaster to reuse successful choices, avoid repeated failures, and spend limited budget on more promising directions.

\subsection{DataTree}

The DataTree organizes DataMaster's search over executable data states by representing alternative data-side decisions as branches of a tree. It allows the agent to explore multiple ways of improving the data from shared starting contexts, while preserving the branch-local state and feedback needed for later decisions. Formally, we denote the DataTree as $\mathcal{T}=(\mathcal{V},\mathcal{E})$, where node $v_0 \in \mathcal{V}$ is the root node associated with the initial data state $D_0$, $\mathcal{V}$ is the set of executed data-engineering nodes, and $\mathcal{E}$ is the set of parent-to-child edges. Each node represents one data-engineering step executed by the agent under the current branch context. Each edge represents parent-to-child context inheritance, allowing the child node to refer to the parent's accumulated data state, artifacts, decisions, and downstream feedback when making the next data-side decision. Specifically, DataTree contains two node types: \emph{red nodes} broaden the search by discovering new external datasets, while \emph{black nodes} deepen the search by exploiting available candidates to construct stronger executable training datasets for the downstream task and fixed algorithm.  

\paragraph{Red Nodes: Data Exploration} \   Red nodes are responsible for expanding the shared Data Pool by discovering external datasets that may be useful for the downstream task. Conditioned on the current search context, the shared Data Pool $\mathcal{P}$, and the retrieved memory $\mathcal{M}_v$ from Global Memory $\mathcal{M}$ for node $v$, a red node performs open-ended external search and lightweight screening, and outputs a set of newly discovered dataset entries $\Delta \mathcal{P}_v$ that are appended to $\mathcal{P}$ for later exploitation by black nodes. In practice, a red node first generates search queries from the task description, target metric, data schema, failure cases, and prior memory. It then screens returned sources for relevance and usability (e.g., metadata, schema compatibility, modality, scale, label structure, and sample previews), while filtering duplicate or benchmark-restricted sources. Each retained entry stores a compact manifest, including its local path or source pointer, format, schema summary, task relevance, and screening notes. This design turns open-world external search into a reusable candidate layer, separating data acquisition from downstream refinement and allowing later nodes to build on prior discoveries rather than restart from scratch.

\paragraph{Black Nodes: Data Exploitation} \   Black nodes exploit the shared Data Pool $\mathcal{P}$ to improve the current training-data pipeline. Conditioned on the available candidate datasets in $\mathcal{P}$ and the retrieved context $\mathcal{M}_v$ for node $v$ from Global Memory $\mathcal{M}$, a black node selects useful sources, adjusts their composition, applies cleaning and transformation rules, constructs features, and adapts the inputs to the fixed downstream training interface. For a black node $v$, these decisions are materialized as an executable \texttt{DataLoader} and associated data-processing logic, yielding the node-specific data state $D_v$, i.e., the executable data configuration produced by node $v$ under the fixed downstream algorithm. The state is then submitted to downstream training through a fixed \texttt{submit} interface, which instantiates the downstream evaluation $E(T,a_0,D_v)$ under the fixed algorithm $a_0$. The resulting score $y_v$ and diagnostics $\phi_v$ are written back into $\mathcal{M}$. In this way, black nodes convert candidate datasets into executable training hypotheses and use downstream training to compare, diagnose, and refine data-side decisions under a fixed algorithm, rather than relying on heuristic data selection alone.
\subsection{Global Memory \& Data Pool: Shared Search State.}

\paragraph{Global Memory} \  The Global Memory serves as the persistent record layer of DataTree, maintaining a one-to-one record for each executed node to trace branch-level search history, node outputs, downstream outcomes, and reusable findings. Concretely, $\mathcal{M}$ stores node-level records of the form
{\setlength{\abovedisplayskip}{3pt}
\setlength{\belowdisplayskip}{2pt}
\begin{equation*}
r_v =
\begin{cases}
(\texttt{red},\; \Delta \mathcal{P}_v,\; \phi_v), & \text{if } v \text{ is a red node},\\
(\texttt{black},\; D_v,\; y_v,\; \phi_v), & \text{if } v \text{ is a black node}.
\end{cases}
\label{eq:global_memory}
\end{equation*}}
For a node $v$, the default retrieved context is inherited from its parent and sibling nodes:
$\mathcal{M}_v = \{r_{Par(v)}\} \cup \{r_u \mid u \in \mathrm{Sib}(v)\}$,
where $Par(v)$ denotes the parent of node $v$ and $\mathrm{Sib}(v)$ denotes the set of its siblings. In addition to this default retrieved context, agents may actively query other red or black node records during execution and may also write newly discovered findings back into $\mathcal{M}$. This makes the search tree coordinated rather than independent: red nodes can avoid redundant discovery and target uncovered directions, while black nodes can reuse validated states, transfer successful processing choices, and use limited training budget more efficiently.

\paragraph{Data Pool} \  
The Data Pool serves as the shared candidate-data layer of DataMaster, turning external sources discovered by red nodes into reusable candidates for later black-node refinement. Each retained entry stores the information needed for downstream use, including the local path, source description, format, schema, metadata, scale, modality, task relevance, and lightweight screening results. Red nodes append newly discovered entries $\Delta \mathcal{P}_v$ into $\mathcal{P}$, and black nodes read from $\mathcal{P}$ to select, compose, and adapt candidate sources into executable training data for the downstream task and fixed algorithm. While the DataTree maintains branch-local search structure and Global Memory stores node-level outcomes, the Data Pool keeps discovered data sources reusable across branches.

\subsection{Tree Growth and Scheduling under Limited Budget}
\label{subsec:growth_scheduling}
With a limited evaluation budget, DataMaster cannot enumerate all reachable executable data states in $\Pi_B(D_0)$, since each candidate state may require downstream training and evaluation. 
We therefore use two budget-control mechanisms: a \emph{growth policy} that decides whether to expand the tree with a red node or a batch of black nodes, and a \emph{scheduling policy} that prioritizes which unexecuted node should consume the next evaluation budget.

\paragraph{Growth policy} \ Tree growth in DataMaster is dynamic rather than predetermined. A red node expands the candidate data space, but its output is only a set of potential datasets; whether these candidates are actually useful can only be determined after they are selected, adapted, and tested by black nodes. For this reason, each red node is followed by a small batch of black nodes that run in parallel and explore different exploitation hypotheses over the newly expanded Data Pool. After a black node is evaluated, the system decides whether the current branch should continue with further black-node refinement or open a new red node for broader data exploration. This decision is made by a lightweight external controller using the latest training feedback and the current branch context. The advantage of this design is that tree growth remains adaptive: it can deepen when the current data configuration still has room for improvement, and broaden when the branch appears under-covered or its current candidates have been largely exhausted.

\paragraph{Scheduling policy} \ While Eq.~\ref{eq:objective} defines the final data state to be found, scheduling specifies how to allocate the next unit of evaluation budget among pending nodes. The scheduling policy provides this online allocation rule: among pending unexecuted nodes, it selects the next node whose execution is estimated to be most useful for approaching a high-performing data state. Under a limited training budget, DataMaster cannot execute all pending nodes and therefore uses a greedy scheduler to decide which one should consume the next unit of budget. The next executed node is
$v_{\mathrm{next}}=\arg\max_{v\in\mathcal{N}}\mathrm{Score}(v)$,
where $\mathcal{N}$ is the set of nodes that have been generated by the growth policy but not yet executed, and the score balances branch quality against insufficient exploration. In our implementation, we instantiate it with a rule inspired by Upper Confidence Bound (UCB)~\citep{auer2002finite,kocsis2006bandit},
{\setlength{\abovedisplayskip}{3pt}
\setlength{\belowdisplayskip}{2pt}
\begin{equation}
\mathrm{Score}(v)
=
\frac{R_v}{N_v}
+
c_t\sqrt{\dfrac{\log N_{Par(v)}}{N_v}},
\label{eq:score}
\end{equation}}
where $R_v$ and $N_v$ are the cumulative reward and visit count associated with the branch rooted at node $v$, $N_{Par(v)}$ is the corresponding count for its parent, and $c_t$ is a decaying exploration coefficient. Black-node evaluation scores are propagated upward to update these branch statistics, so that repeatedly successful branches become easier to prioritize while still retaining an incentive to explore insufficiently tested directions. Additional implementation details, including reward backpropagation, initialization of newly generated nodes, and the decay schedule of $c_t$, are provided in Appendix~\ref{app:scheduling_details}.

\paragraph{Overall} DataMaster instantiates autonomous data engineering as a closed-loop search system over executable data configurations. DataTree organizes alternative data-side decisions, Data Pool shares discovered external sources, and Global Memory reuses prior outcomes so that data discovery, refinement, and downstream validation reinforce each other rather than remain isolated steps. Fig.~\ref{fig:datamaster_random_acts_walkthrough} shows an example of the whole system. The following experiments evaluate whether this structured data-side search can improve fixed algorithms across both classical machine learning and LLM post-training settings.

\section{Experiments}
\label{sec:experiments}

\begin{table*}[t]
    \centering
    \caption{Main result comparison on MLE-Bench Lite and PostTrainBench. Unless marked otherwise, agentic systems use GLM-5; $^*$Codex is driven by GPT-5.2-Codex. Best baseline scores are in \textbf{bold}.}
    \label{tab:main}
    \setlength{\tabcolsep}{5pt}
    \renewcommand{\arraystretch}{1.15}
    \resizebox{\textwidth}{!}{%
    \begin{tabular}{l|cc|cccccccc}
    \toprule
     & \multicolumn{2}{c|}{\textbf{MLE-Bench Lite}} & \multicolumn{8}{c}{\textbf{PostTrainBench}} \\
    \cmidrule(lr){2-3}\cmidrule(lr){4-11}
    \textbf{System} 
    & \makecell{Medal\\Rate} 
    & \makecell{Gold Medal\\Rate} 
    & \makecell{AIME\\25} 
    & \makecell{ArenaHard\\Writing} 
    & \makecell{BFCL} 
    & GPQA 
    & GSM8K 
    & \makecell{HealthBench\\Easy} 
    & HumanEval 
    & Avg \\
    \midrule

    \multicolumn{11}{c}{\cellcolor{yellow!12}\textbf{Reference Points}} \\
    Initial Score 
    & 35.91\% 
    & 22.73\% 
    & 0.00\% 
    & 0.48\% 
    & 22.56\% 
    & 18.75\% 
    & 9.02\% 
    & 0.00\% 
    & 8.53\% 
    & \cellcolor{gray!15}8.47\% \\[2pt]

    Human Score 
    & -- 
    & -- 
    & 6.66\% 
    & 44.84\% 
    & 63.46\% 
    & 30.35\% 
    & 86.95\% 
    & 35.76\% 
    & 60.36\% 
    & \cellcolor{gray!15}46.91\% \\

    \midrule
    \multicolumn{11}{c}{\cellcolor{orange!12}\textbf{Baselines}} \\

    Codex* 
    & 22.73\% 
    & 18.18\% 
    & 0.00\% 
    & 0.15\% 
    & \textbf{40.67\%} 
    & 29.54\% 
    & 24.31\% 
    & 9.17\% 
    & 25.41\% 
    & \cellcolor{gray!15}18.46\% \\[2pt]

    Claude Code 
    & 36.36\% 
    & 22.12\% 
    & 0.00\% 
    & 2.01\% 
    & 31.00\% 
    & 14.29\% 
    & 2.58\% 
    & 1.16\% 
    & 29.27\% 
    & \cellcolor{gray!15}11.47\% \\[2pt]

    DatasetResearcher 
    & 59.09\% 
    & 27.27\% 
    & 0.00\% 
    & 0.00\% 
    & 22.64\% 
    & 18.00\% 
    & 9.33\% 
    & 3.56\% 
    & 6.00\% 
    & \cellcolor{gray!15}8.50\% \\[2pt]

    DataFlex 
    & -- 
    & -- 
    & 0.00\% 
    & 2.07\% 
    & 22.58\% 
    & 28.80\% 
    & 48.37\% 
    & 6.22\% 
    & 25.61\% 
    & \cellcolor{gray!15}19.09\% \\[1pt]

    ML-Master 2.0 
    & 40.91\% 
    & 27.27\% 
    & 3.33\% 
    & 0.00\% 
    & 31.43\% 
    & 18.57\% 
    & 31.45\% 
    & 8.24\% 
    & 16.84\% 
    & \cellcolor{gray!15}15.69\% \\

    \midrule

    \rowcolor{orange!25}
    \textbf{DataMaster} 
    & \textbf{68.18\%} 
    & \textbf{45.45\%} 
    & \textbf{3.33\%} 
    & \textbf{21.93\%} 
    & 34.27\% 
    & \textbf{31.02\%} 
    & \textbf{49.43\%} 
    & \textbf{34.93\%} 
    & \textbf{43.29\%} 
    & \cellcolor{gray!15}\textbf{31.17\%} \\

    \rowcolor{orange!25}
    & {\tiny\color{green!60!black}(+32.27\%)} 
    & {\tiny\color{green!60!black}(+22.72\%)} 
    & 
    & 
    & 
    & 
    & 
    & 
    & 
    & \cellcolor{gray!15}{\tiny\color{green!60!black}(+22.70\%)} \\

    \bottomrule
    \end{tabular}%
    }
\end{table*}

\subsection{Setup}

\paragraph{Benchmarks} \  
We evaluate DataMaster on two complementary benchmarks that differ in their data-availability assumptions.
\textbf{MLE-Bench Lite}~\citep{chan2025mlebenchevaluatingmachinelearning} is a held-out subset of MLE-Bench spanning diverse Kaggle competitions across tabular, vision, NLP, and time-series tasks. Each task provides a fixed competition dataset; the agent must improve downstream performance solely through data-side operations (selection, cleaning, feature engineering) while leaving the model training code unchanged.
\textbf{PostTrainBench}~\citep{rank2026posttrainbenchllmagentsautomate} measures how well AI agents can post-train a base language model given only compute and a time budget, with no training data provided. The agent must discover and curate datasets from scratch, covering seven capabilities (details in Appendix~\ref{app:benchmark_details}).

\paragraph{Metrics} \  
On \emph{MLE-Bench Lite} we report four metrics: (i)~\textbf{Medal Rate}, the fraction of competitions where the agent's best node achieves a medal-level submission; (ii)~\textbf{Gold Medal Rate}, restricted to gold medals; (iii)~\textbf{Avg.\ Overcome Rate}, the fraction of nodes in DataTree whose submission score exceeds the initial-node submission score; and (iv)~\textbf{Performance Gain}, the mean relative improvement of the best score in DataTree over the initial submission.
On \emph{PostTrainBench} we report per-task accuracy for each of the seven capabilities above together with the macro-average (\textbf{Avg}).

\paragraph{Baselines} \ 
We compare DataMaster against four categories of baselines: \textbf{(i) Frontier agents}, including Claude Code CLI~\citep{anthropic2024claudecode} and Codex CLI~\citep{openai2025codexcli}; \textbf{(ii) Dataset Search Agent}, Dataset Research~\citep{li2025datasetresearchbenchmarkingagentsystems}; \textbf{(iii) End-to-End ML Agent}, ML-Master-2~\citep{zhu2026ultralonghorizonagenticsciencecognitive}; and \textbf{(iv) Data-Centric Training Framework}, DataFlex~\citep{liang2026dataflex}.
\textbf{Reference Points}: On MLE-Bench Lite, the initial score is obtained by running ML-Master's code with its data module removed; on PostTrainBench, it corresponds to the base model (Qwen3-1.7B-Base) without fine-tuning. \textbf{Human score} refers to the expert-trained instruct model (Qwen3-1.7B), serving as a strong human-curated reference for assessing whether autonomous data engineering can approach expert-designed post-training. See detailed implementations in Appendix~\ref{app:baseline_details}.
\subsection{Main Results}

Table~\ref{tab:main} presents the main results comparing DataMaster against all baselines on MLE-Bench Lite and PostTrainBench. All systems are powered by GLM-5~\citep{glm5team2026glm5vibecodingagentic} as the backbone LLM, with a 12-hour time limit and identical training compute budget per task.

\paragraph{Comparison with Initial Score} \  
As shown in Table~\ref{tab:main}, DataMaster substantially outperforms the initial score on both benchmarks.
On MLE-Bench Lite, it achieves a medal rate of \textbf{68.18\%} and a gold medal rate of \textbf{45.45\%}, demonstrating that operating \emph{exclusively} on data while leaving the training code untouched can deliver substantial downstream gains.
On PostTrainBench, DataMaster raises the average score from 8.47\% to 31.17\% (+22.70 percentage points), with particularly large improvements on GSM8K (9.02\%$\rightarrow$49.43\%), HealthBench Easy (0.00\%$\rightarrow$34.93\%), and HumanEval (8.53\%$\rightarrow$43.29\%).

\paragraph{Comparison with Other Baselines} \  
DataMaster consistently outperforms all baselines on both benchmarks.
On MLE-Bench Lite, it achieves the highest medal rate (68.18\%) and gold medal rate (45.45\%), surpassing DatasetResearch by 9.09 percentage points in medal rate and 18.18 percentage points in gold medal rate, indicating that coupling external data discovery with iterative refinement is more effective than search alone.
On PostTrainBench, DataMaster achieves the highest average score (31.17\%) among all systems. Compared to Codex (18.46\%), which has strong single-task performance on BFCL (40.67\%) but struggles with long-horizon refinement, DataMaster achieves more balanced improvements across all seven capabilities. Compared to DataFlex (19.09\%), which receives the same curated dataset but only applies training-time optimization, DataMaster outperforms by 12.08 percentage points, demonstrating that data curation quality matters more than training-time scheduling alone. Notably, DataMaster also surpasses the human-expert instruct model on GPQA (31.02\% vs. 30.35\%), suggesting that autonomous data engineering can approach or exceed expert-curated post-training data on selected capabilities.

\subsection{Ablation Study}
\label{sec:ablation}

\begin{figure}[ht]
    \centering
    \includegraphics[width=0.92\linewidth]{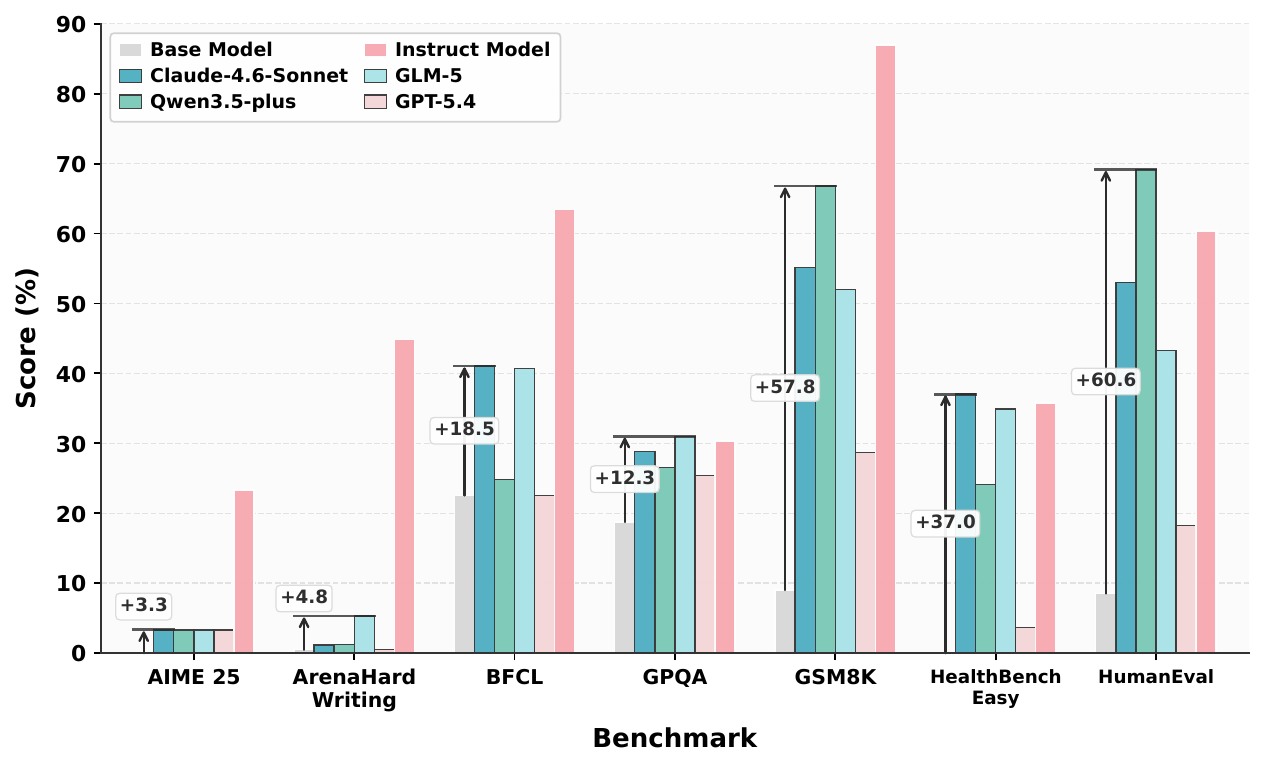}
    \caption{PostTrainBench performance across LLMs. Per-task accuracy of DataMaster powered by four frontier LLMs, compared against Base Model and Instruct Model baselines.}
    \label{fig:benchmark_comparison}
\end{figure}

\paragraph{Performance Across Different LLMs} \quad
To understand how DataMaster varies with backbone LLMs, we evaluate GLM-5, Claude-4.6-sonnet~\citep{anthropic2026sonnet46}, Qwen3.5-Plus~\citep{qwen3.5}, and GPT-5.4~\citep{openai2026gpt54} on PostTrainBench. Fig.~\ref{fig:benchmark_comparison} shows per-task accuracy across seven capabilities. Key findings:
\ding{182}~GLM-5, Claude-4.6-sonnet, and Qwen3.5-plus all achieve around 30\% average accuracy, substantially above the Base Model (8.47\%), showing that DataMaster is effective across multiple strong backbones.
\ding{183}~DataMaster reaches or surpasses the human-expert instruct model on selected capabilities: GLM-5 on GPQA, Claude-4.6-sonnet on HealthBench Easy, and Qwen3.5-plus on HumanEval.
\ding{184}~GPT-5.4 performs worse (14.64\% average). We observe that it makes far fewer tool calls per node (11.3 on average) than other backbones (40.7 for GLM-5), suggesting that insufficient tool use may limit long-horizon data search and refinement.

\begin{figure}[ht]
    \centering
    \includegraphics[width=0.82\linewidth]{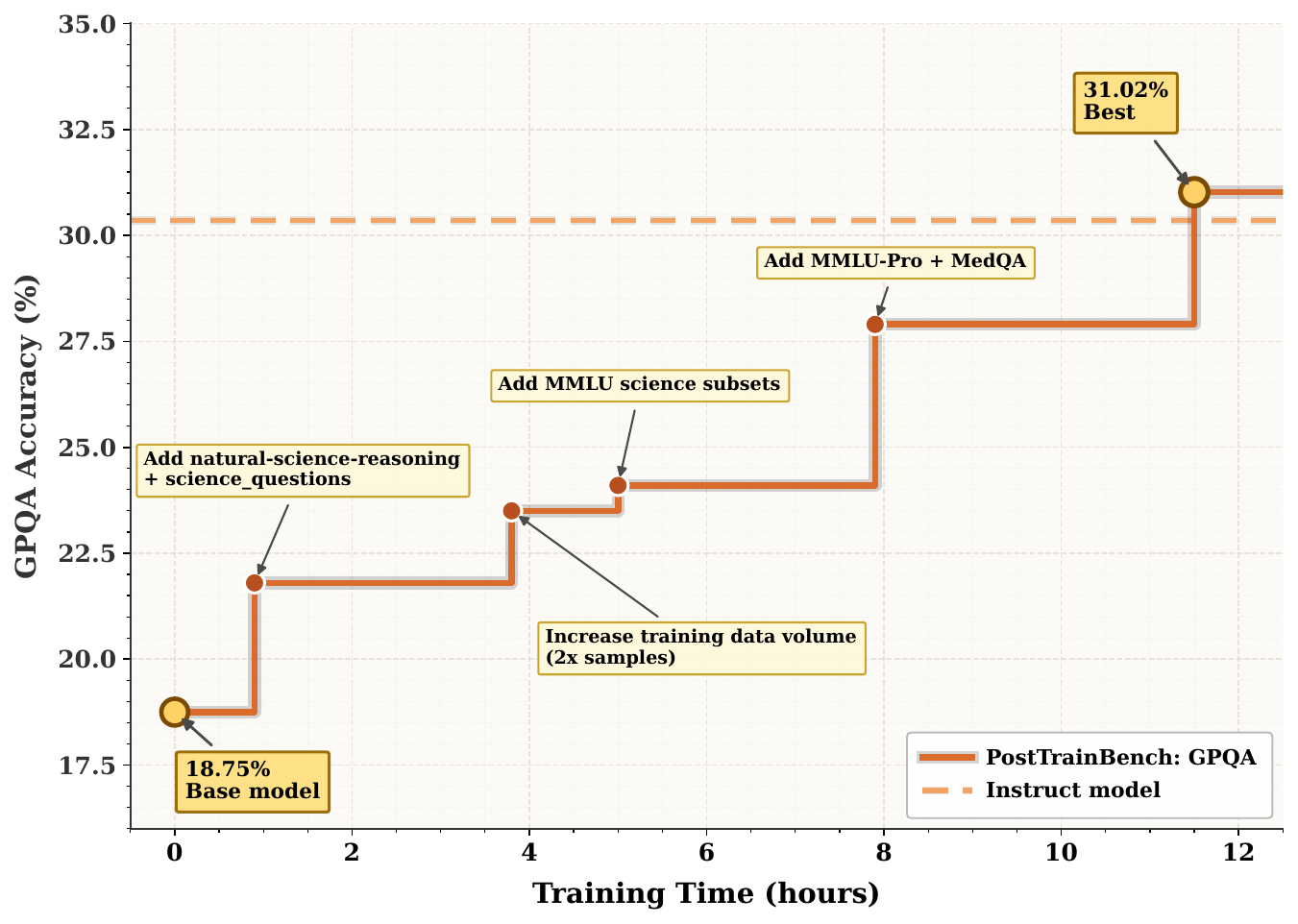}
    \caption{Test-time scaling on PostTrainBench. Best-node score as a function of wall-clock time, showing continuous improvement as budget grows.}
    \label{fig:test_time_scaling_posttrain}
\end{figure}

\paragraph{Test-Time Scaling} \quad
We examine test-time scaling on the GPQA task in PostTrainBench by tracking the best accuracy over wall-clock training time. As shown in Fig.~\ref{fig:test_time_scaling_posttrain}, DataMaster improves stepwise from the base score of 18.75\% to 31.02\% by progressively discovering and integrating more relevant science, reasoning and MedQA~\citep{jin2021disease} data. The final node surpasses the instruct model reference, showing that additional search budget can produce stronger task-specific data configurations rather than merely longer execution.

\paragraph{Component Analysis} \quad
We ablate DataMaster's core components: \emph{red} nodes (data exploration), \emph{black} nodes (data exploitation), and \emph{Global Memory} (cross-iteration knowledge store), on 10 MLE-Bench Lite tasks that cover the main task types in the benchmark, including tabular, vision, NLP, time-series, and audio tasks. Each variant disables one component while keeping the other two active. Table~\ref{tab:component_analysis} shows initial-node and best-node metrics. We find that: \textbf{\ding{182}~Memory enables cross-iteration learning.}
Without memory, overcome rate drops from 69.96\% to 28.57\% and best gold medal rate from 80\% to 20\%, negating all gains.
\textbf{\ding{183}~Red and black nodes are complementary.}
Without red node, internal refinement improves (overcome rate 72.40\%), but best gold medal rate drops 10\%, showing external discovery adds unique value. Without black node, discovered data cannot be adapted (lower medal rate), showing external sources need in-place refinement.
\textbf{\ding{184}~Full system achieves strongest performance.}
Only with all three components enabled does DataMaster reach a 90\% medal rate.

\begin{table}[ht]
\centering
\caption{%
  Component analysis on MLE-Bench Lite.
  \cmark/\xmark{} indicates whether the component is enabled.
  \emph{Initial-code} metrics measure the quality of the initial solution node;
  \emph{best-node} metrics reflect the best submission found across the full search tree.
}
\label{tab:component_analysis}
\renewcommand{\arraystretch}{1.05}
\resizebox{0.8\linewidth}{!}{
\begin{tabular}{ccc @{\hspace{14pt}} cc @{\hspace{14pt}} cc @{\hspace{14pt}} c}
\toprule
\multirow{2}{*}{\textbf{Red}} &
\multirow{2}{*}{\textbf{Black}} &
\multirow{2}{*}{\textbf{Memory}} &
\multicolumn{2}{c}{\textbf{Initial-node}} &
\multicolumn{2}{c}{\textbf{Best-node}} &
\multirow{2}{*}{\makecell{\textbf{Overcome} \\ \textbf{Rate}}} \\
\cmidrule(lr){4-5}\cmidrule(lr){6-7}
& & &
\makecell{Medal Rate} & \makecell{Gold\\Medal Rate} &
\makecell{Medal Rate} & \makecell{Gold\\Medal Rate} & \\
\midrule
\xmark & \cmark & \cmark & 50\% & 20\% & 80\% {\scriptsize\color{green!60!black}(+30\%)} & 70\% {\scriptsize\color{green!60!black}(+50\%)} & \textbf{72.40}\% \\
\cmark & \xmark & \cmark & 60\% & 20\% & 70\% {\scriptsize\color{green!60!black}(+10\%)} & 50\% {\scriptsize\color{green!60!black}(+30\%)} & 52.41\% \\
\cmark & \cmark & \xmark & 50\% & 20\% & 50\% {\scriptsize\color{green!60!black}(+0\%)} & 20\% {\scriptsize\color{green!60!black}(+0\%)} & 28.57\% \\
\midrule
\cmark & \cmark & \cmark & 50\% & 30\% & \textbf{90}\% {\scriptsize\color{green!60!black}(+\textbf{40}\%)} & \textbf{80}\% {\scriptsize\color{green!60!black}(+\textbf{50}\%)} & 69.96\% \\
\bottomrule
\end{tabular}}
\end{table}

\paragraph{Robustness to Algorithm Initialization} \quad 
A practical question is whether DataMaster depends on a specific fixed algorithm implementation. Each MLE-Bench Lite task provides an initial solution with a \emph{DataLoader} for data loading, preprocessing, and augmentation, and an \emph{Algorithm} for model construction, training, and inference (Appendix~\ref{sec:init_code_examples}). We compare \textbf{Full-code}, where both parts are provided, with \textbf{Algo-only}, where only the Algorithm skeleton is given and the agent must construct the DataLoader.
Table~\ref{tab:init_code} shows that:
\textbf{\ding{182}}~DataMaster improves over the initial baseline under both settings, showing that data-side optimization is robust to different fixed-algorithm starting points.
\textbf{\ding{183}}~Algo-only starts from a lower initial medal rate (35.91\% vs.\ 45.45\%) yet achieves a larger absolute gain (+32.27\% vs.\ +27.28\%) and a higher overcome rate (67.31\% vs.\ 60.40\%), indicating that DataMaster can compensate for a weaker starting point by constructing better data pipelines.

\begin{table}[ht]
\centering
\caption{%
  Sensitivity to initial code quality on MLE-Bench Lite.
  \emph{Full-code} starts from a complete training pipeline;
  \emph{Algo-only} starts from a minimal algorithmic skeleton.
  All runs use DataMaster (full) with all three components enabled.
  Green annotations indicate the absolute gain over the corresponding initial-code metric.
}
\label{tab:init_code}
\renewcommand{\arraystretch}{1.05}
\resizebox{0.8\linewidth}{!}{
\begin{tabular}{l @{\hspace{14pt}} cc @{\hspace{14pt}} cc @{\hspace{14pt}} c}
\toprule
\multirow{2}{*}{\textbf{Initial Code}} &
\multicolumn{2}{c}{\textbf{Initial-node}} &
\multicolumn{2}{c}{\textbf{Best-node}} &
\multirow{2}{*}{\makecell{\textbf{Overcome}\\\textbf{Rate}}} \\
\cmidrule(lr){2-3}\cmidrule(lr){4-5}
& \makecell{Medal Rate} & \makecell{Gold\\Medal Rate}
& \makecell{Medal Rate} & \makecell{Gold\\Medal Rate} & \\
\midrule
Full-code  & 45.45\% & 18.18\% & \textbf{72.73}\% {\scriptsize\color{green!60!black}(+27.28\%)} & 45.45\% {\scriptsize\color{green!60!black}(+\textbf{27.27}\%)} & 60.40\% \\
Algo-only  & 35.91\% & 22.73\% & 68.18\% {\scriptsize\color{green!60!black}(+\textbf{32.27}\%)} & \textbf{45.45}\% {\scriptsize\color{green!60!black}(+22.72\%)} & \textbf{67.31}\% \\
\bottomrule
\end{tabular}}
\end{table}

\subsection{Analysis}
\label{sec:analysis}

\textbf{Cost Analysis} \quad 
We analyze DataMaster's internal cost distribution on MLE-Bench Lite, reporting per-node token usage, API cost, tool calls, and wall-clock time for red and black nodes. Table~\ref{tab:node_cost} summarizes the average costs over 641 red nodes and 340 black nodes.
We find that: \ding{182}~Black nodes are costlier, using 3.0$\times$ more API budget and taking 2.4$\times$ longer than red nodes, due to data selection, processing, DataLoader construction, and downstream training/evaluation. \ding{183}~Red nodes dominate search breadth, with nearly twice as many executions as black nodes, reflecting open-ended external data discovery. \ding{184}~This cost profile matches DataMaster's design: red nodes cheaply expand the candidate space, while black nodes spend more budget validating downstream gains.

\begin{table}[ht]
\centering
\caption{%
  Node-wise cost breakdown within DataMaster's search tree on MLE-Bench Lite.}
\label{tab:node_cost}
\renewcommand{\arraystretch}{1.12}
\scalebox{0.8}{
\begin{tabular}{lccc}
\toprule
& \multicolumn{2}{c}{\textbf{Node Type}} & \\
\cline{2-3}
\textbf{Metric}
& \cellcolor{red!15}\textcolor{red!70!black}{\textbf{Red}}
& \cellcolor{black!12}\textbf{Black}
& \textbf{All} \\
\midrule
\# Nodes & 641 & 340 & 981 \\
\midrule
Input tokens (K)  & 55.2 & 68.1 & 59.4 \\
Output tokens (K) & 6.9  & 24.9 & 13.2 \\
\midrule
API cost (\$)     & 0.021 & 0.062 & 0.036 \\
Avg. tool calls   & 41.3  & 39.6  & 40.7 \\
\midrule
Tool time (min)   & 16.7 & 38.5 & 24.0 \\
Total time (min)  & 21.3 & 50.6 & 31.6 \\
\bottomrule
\end{tabular}}
\end{table}

\begin{figure}[ht]
\centering
\begin{minipage}[t]{0.47\linewidth}
\centering
\includegraphics[width=\linewidth]{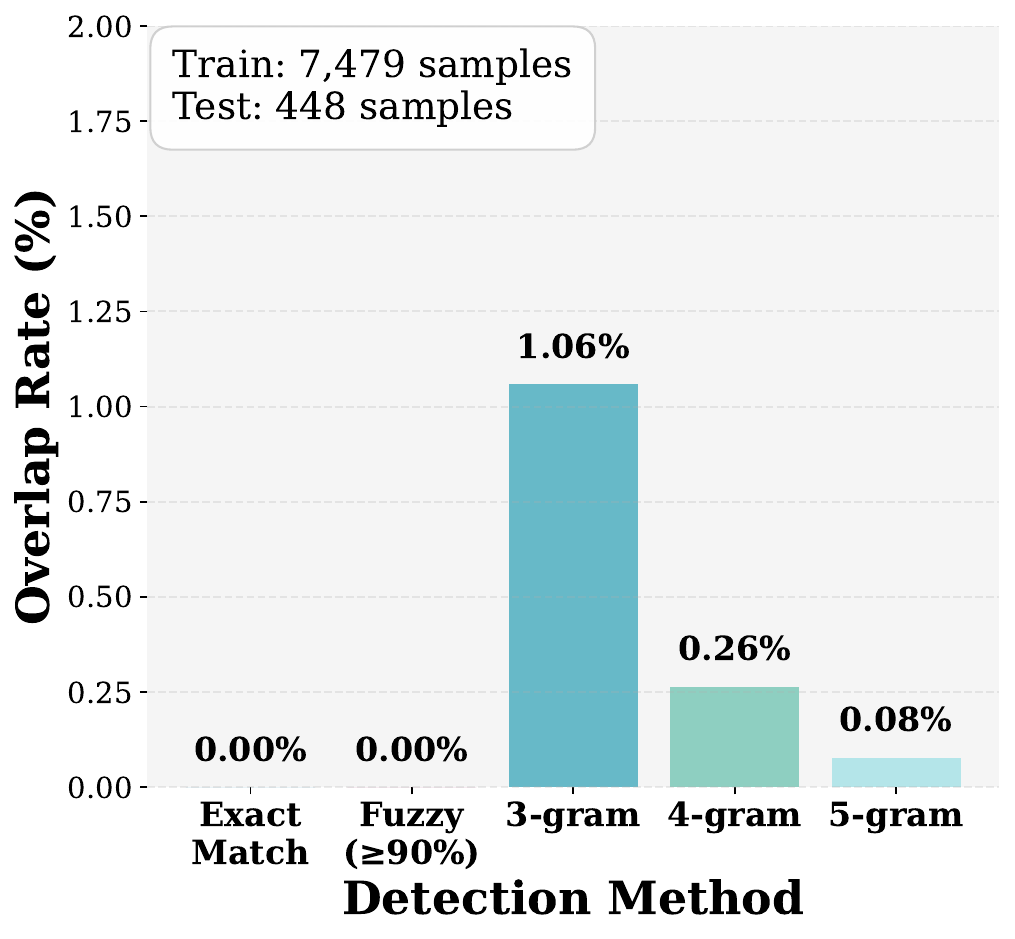}
\caption{Test-set leakage check. Test-train overlap on PostTrainBench GPQA.}
\label{fig:data_leakage}
\end{minipage}
\hfill
\begin{minipage}[t]{0.47\linewidth}
\centering
\includegraphics[width=\linewidth]{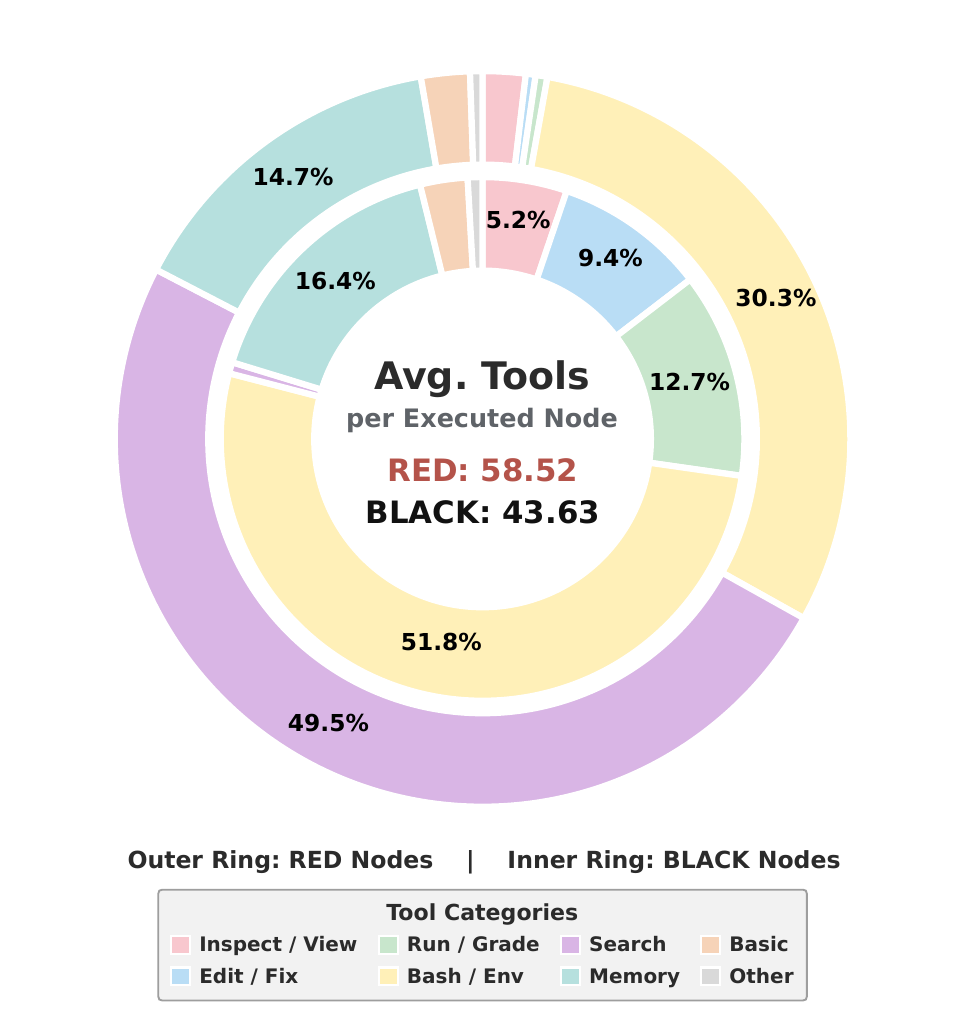}
\caption{Tool-use composition. Tool-use statistics of Red/Black nodes.}
\label{fig:red_black_tool_donut}
\end{minipage}
\end{figure}

\noindent \textbf{Data Leakage Analysis} \quad 
\label{sec:leakage}
A concern in open-ended data search is test-set contamination. We address this through a multi-layer protocol: \ding{182}~\textbf{source filtering} that blocklists benchmark pages, competition discussions, and URLs matching test-split names, preventing test data from entering the training pipeline; \ding{183}~\textbf{hash-based deduplication} that stores test-set hashes and filters out any training samples with exact matches; \ding{184}~\textbf{provenance tracking} that records URL, timestamp, and content hash for every external data point, enabling leakage audits. 
Fig.~\ref{fig:data_leakage} audits GPQA leakage over 7,479 discovered training samples. We find zero exact or fuzzy matches and only low 3--5-gram overlap (0.08--1.06\%), suggesting terminology overlap rather than test-set copying.

\paragraph{Statistic Analysis} \quad 
We analyze the tool-use composition of red and black nodes in Fig.~\ref{fig:red_black_tool_donut}.
Key findings:
\ding{182}~red nodes use more tools per execution than black nodes on average
(58.52 vs.\ 43.63), reflecting the interaction-heavy nature of external data discovery.
\ding{183}~Red-node actions are dominated by Search (49.5\%), followed by Bash/Env
(30.3\%) and Memory (14.7\%), showing that red nodes primarily explore candidate
data sources, inspect artifacts, and reuse prior search context.
\ding{184}~Black-node actions are dominated by Bash/Env operations (51.8\%), with
non-trivial Run/Grade, Edit/Fix, and Inspect/View usage, indicating that black nodes
focus on constructing executable data states and validating them through downstream
execution rather than broad source discovery.

\section{Conclusion}
\label{sec:conclusion}

We introduced task-conditioned autonomous data engineering as a new problem where an agent improves a fixed training algorithm by searching over data states. We propose DataMaster to solve this problem through a DataTree for branching data-side search, a shared Data Pool for reusable external sources, and a Global Memory for carrying outcomes, artifacts, and reusable findings across iterations. Experiments on MLE-Bench Lite and PostTrainBench show that DataMaster consistently improves strong initial solutions and outperforms agentic and data-centric baselines, demonstrating that data-side search can be an effective optimization lever beyond model-code modification. These results suggest that future autonomous ML systems should place greater emphasis on data-side improvement, including how data is discovered, selected, transformed, validated, and reused across tasks.

\bibliography{main}


\appendix

\clearpage

\section{Case Study: DataMaster on Random Acts of Pizza}

To provide a concrete view of the DataMaster workflow, we include a representative walkthrough on the \texttt{random-acts-of-pizza} task. This task requires predicting whether a Reddit pizza request will receive a successful response, using request text and associated metadata. The figure illustrates how DataMaster organizes data engineering as a tree-structured search process, where different branches correspond to alternative data-side decisions under the same fixed algorithm.

\begin{figure}[htbp]
    \centering
    \includegraphics[width=0.98\linewidth]{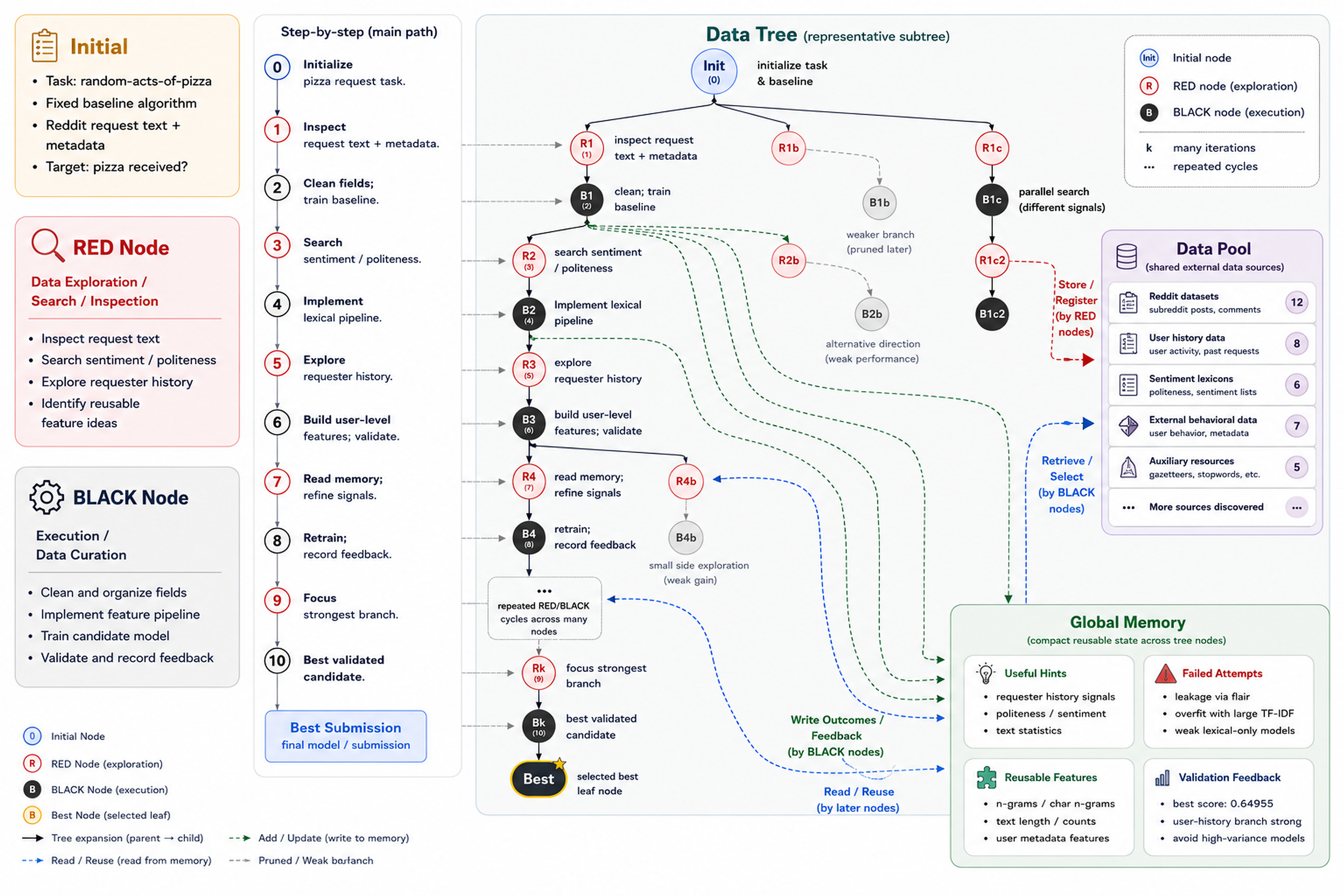}
    \caption{
    Case walkthrough of DataMaster on the \texttt{random-acts-of-pizza} task.
    }
    \label{fig:datamaster_random_acts_walkthrough}
\end{figure}

This walkthrough highlights the separation between the three core components. The DataTree provides the tree-based search structure, allowing the agent to explore multiple data-engineering directions. The Data Pool serves as the shared candidate-data layer: Red nodes write newly discovered external data entries into it, and Black nodes later read from it to construct executable training inputs. Global Memory is a separate record layer that stores node outcomes, failed trials, useful hints, and validation feedback for later reuse. Together, these components allow DataMaster to broaden the search over external data sources, refine selected candidates through downstream validation, and carry reusable information across branches without conflating discovered data sources with search memory.

\section{Experiment Details}
\label{app:benchmark_details}
\subsection{Set up}
\paragraph{Hardware configuration.}
We run all MLE-Bench Lite experiments on an 8-GPU machine with NVIDIA H20 GPUs. To enable parallel execution, we allocate one H20 GPU per 4 tasks, allowing multiple competitions to run concurrently while sharing compute resources efficiently.

\subsection{Benchmark Details}
\subsubsection{MLE-Bench Lite}

MLE-Bench Lite~\citep{chan2025mlebenchevaluatingmachinelearning} is a curated subset of MLE-Bench containing 22 Kaggle competitions spanning tabular prediction, computer vision, natural language processing, and time-series forecasting. Each task provides a fixed competition dataset and an initial code solution. The agent operates exclusively on the data side—selecting subsets, engineering features, cleaning outliers, and adjusting data formatting—while the model architecture and training script remain unchanged. Performance is measured by whether the agent's best submission achieves a Kaggle medal threshold (bronze, silver, or gold), with medal rate and gold medal rate as the primary metrics.

\subsubsection{PostTrainBench}

PostTrainBench~\citep{rank2026posttrainbenchllmagentsautomate} evaluates autonomous data engineering for post-training a base language model. Unlike MLE-Bench Lite, \emph{no training data is provided}; the agent must discover, collect, and curate all datasets from scratch. The benchmark measures downstream accuracy after fine-tuning the base model on the agent-curated data. It covers seven diverse capabilities:

\begin{itemize}
    \item \textbf{AIME 2025}~\citep{aime2025} — mathematical reasoning, evaluated on competition-level math problems.
    \item \textbf{Arena-Hard Writing}~\citep{arenahard2024} — instruction following and creative writing, judged by an LLM judge.
    \item \textbf{BFCL}~\citep{patil2025bfcl} — function calling, testing the model's ability to generate correct API calls.
    \item \textbf{GPQA}~\citep{rein2024gpqa} — graduate-level scientific knowledge across physics, chemistry, and biology.
    \item \textbf{GSM8K}~\citep{cobbe2021gsm8k} — grade-school arithmetic reasoning with multi-step word problems.
    \item \textbf{HealthBench Easy}~\citep{aroraHealthBenchEvaluatingLarge2025} — medical question answering on clinical scenarios.
    \item \textbf{HumanEval}~\citep{chen2021humaneval} — code generation, measuring functional correctness of synthesized Python programs.
\end{itemize}

The primary metric is per-task accuracy, together with the macro-average across all seven tasks.

\subsubsection{Key Differences Between Benchmarks}

Table~\ref{tab:benchmark_comparison} summarizes the key differences between the two evaluation settings.

\begin{table}[h]
\centering
\caption{Comparison of evaluation settings.}
\label{tab:benchmark_comparison}
\begin{tabular}{lcc}
\toprule
& \textbf{MLE-Bench Lite} & \textbf{PostTrainBench} \\
\midrule
Initial data provided & Yes (competition dataset) & No \\
Data discovery required & No & Yes (open-world search) \\
Task domains & Kaggle ML competitions & LLM post-training capabilities \\
Number of tasks & 22 & 7 \\
Fixed component & Training code & Training script + base model \\
Primary metric & Medal rate & Per-task accuracy \\
\bottomrule
\end{tabular}
\end{table}

\subsection{Baseline Implementation Details}
\label{app:baseline_details}

\subsubsection{Baseline}

\paragraph{DataFlex} DataFlex~\citep{liang2026dataflex} is a data-centric training framework built on LLaMA-Factory that dynamically controls data selection, mixture ratios, and sample reweighting during training. Unlike DataMaster, it does not perform autonomous dataset discovery; it assumes candidate datasets are already available and optimizes their usage during training.

Because DataFlex is built on LLaMA-Factory, it is only applicable to PostTrainBench (language model post-training) and cannot be evaluated on MLE-Bench Lite (heterogeneous ML tasks). For PostTrainBench evaluation, we provide DataFlex with the same curated dataset produced by DataMaster's best node, isolating DataFlex's training-time optimization from the data discovery process. In principle, DataMaster (data discovery and curation) and DataFlex (training-time optimization) are orthogonal and could be combined.

\paragraph{Claude Code CLI.}
Claude Code CLI~\citep{anthropic2024claudecode} is Anthropic's command-line agent for agentic coding tasks. It has access to file operations, bash execution, and web search, and can iteratively refine code and data pipelines. We evaluate it on both benchmarks by providing the same initial setup as DataMaster.

\paragraph{Codex CLI.}
For MLE-Bench Lite, we use the Codex CLI implementation provided by the benchmark authors. For PostTrainBench, we report results using GPT-5.2-codex from the original PostTrainBench paper~\citep{rank2026posttrainbenchllmagentsautomate}. We do not run our own Codex CLI evaluation on PostTrainBench because: (1) the compact API does not support external model endpoints, and (2) when integrated with GLM-5.1, Codex CLI frequently exceeds the maximum token limit and terminates prematurely, preventing long-horizon iterative refinement.

\paragraph{DatasetResearch.}
DatasetResearch~\citep{li2025datasetresearchbenchmarkingagentsystems} is a pure data-search agent that discovers external datasets but does not perform code adaptation or training. For our evaluation, we run DatasetResearch to discover datasets, then use Claude Code CLI to adapt the discovered data into the training pipeline. To ensure fair comparison, we control the total training data volume to match DataMaster's best node.

\paragraph{ML-Master 2.0.}
ML-Master 2.0~\citep{zhu2026ultralonghorizonagenticsciencecognitive} is an end-to-end machine learning agent that can modify both data and code. We evaluate it on MLE-Bench Lite using the authors' released implementation. It is not evaluated on PostTrainBench due to implementation constraints.

\subsection{Reference Points}

\paragraph{Initial Score.}
The initial score represents the baseline performance before any autonomous data engineering.

\begin{itemize}

\item On \textbf{MLE-Bench Lite}, the initial score is obtained by running ML-Master 2.0's code with its data module removed. This isolates the contribution of the fixed training pipeline (model architecture, hyperparameters, training script) from any data-side improvements. The resulting medal rate (35.91\%) and gold medal rate (22.73\%) serve as the starting point for measuring data engineering gains.

\item On \textbf{PostTrainBench}, the initial score corresponds to the base model \textbf{Qwen3-1.7B-Base} evaluated directly on the seven downstream tasks without any fine-tuning. This represents the zero-shot or few-shot performance of the pretrained model before any task-specific data curation or training. The average score of 8.47\% reflects the base model's limited capability on specialized tasks without post-training.

\end{itemize}

\paragraph{Human Score.}
The human score represents the performance ceiling achieved by human experts through manual data curation and training. On \textbf{PostTrainBench}, the human score corresponds to \textbf{Qwen3-1.7B}, the instruct-tuned version of the base model, trained by Alibaba's expert team on carefully curated instruction-following datasets. The average score of 46.91\% represents the state-of-the-art achievable through expert human data curation and serves as the target for autonomous data engineering systems.

\subsection{Per-Task Performance Analysis}
\label{sec:perf_details}

\begin{figure}[h]
  \centering
  \includegraphics[width=0.6\linewidth]{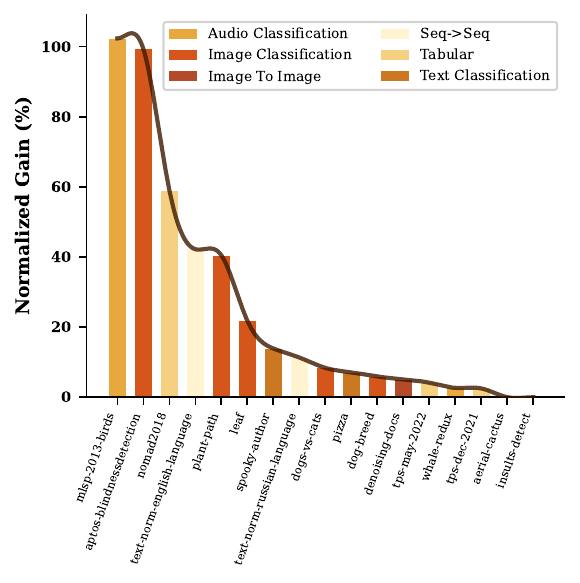}
  \caption{%
    Task-wise normalized gain and overcome rate on 17 MLE-Bench Lite tasks.
    All tasks achieve non-negative gain; top performers span Audio, Tabular, Image Classification, and Seq->Seq domains.
  }
  \label{fig:taskwise_perf}
\end{figure}

We analyze DataMaster's per-task performance on 17 MLE-Bench Lite tasks (5 tasks excluded due to invalid initial submissions). We report \emph{normalized gain}, which measures the best-node improvement relative to the median-to-gold score range, and \emph{overcome rate}, the fraction of DataTree nodes that surpass the initial-node score. Fig.~\ref{fig:taskwise_perf} visualizes both metrics across all tasks.

\textbf{Key findings:}
\ding{182}~DataMaster achieves non-negative gain on all 17 tasks, never degrading performance. Top performers show both high gain and high overcome rate (e.g., 102.33\% gain with 85.71\% overcome on an audio classification task).
\ding{183}~Gain and overcome rate are not always aligned: some tasks exhibit high gain with low overcome rate, where few nodes discover exceptional configurations; others show low gain with high overcome rate, where the initial score is already near-optimal yet DataMaster still improves steadily.
\ding{184}~Performance varies by domain: audio and sequence-to-sequence tasks achieve the highest gains, followed by tabular tasks, while image/text classification tasks show more variable gains---particularly when external data is scarce or train/test distributions are near-identical.

\paragraph{Normalized gain computation.}

For each competition, we compute a \emph{normalized gain} that measures the best-node improvement relative to the score range between the median and gold thresholds:
\begin{equation}
  \text{Gain\%} =
  \begin{cases}
    \dfrac{\text{best} - \text{initial}}{|\text{gold} - \text{median}|} \times 100 & \text{higher-is-better}, \\[8pt]
    \dfrac{\text{initial} - \text{best}}{|\text{gold} - \text{median}|} \times 100 & \text{lower-is-better}.
  \end{cases}
\end{equation}
This normalization accounts for the varying difficulty across competitions: a 0.01 improvement on a tightly clustered metric may represent a larger relative achievement than a 0.05 improvement on a widely spread one.

Table~\ref{tab:perf_details} reports the full per-task results, including the initial-node score, best-node score, relative improvement, normalized gain, and overcome rate.

\definecolor{gainhigh}{rgb}{0.85,0.33,0.10}
\definecolor{gainmed}{rgb}{0.92,0.53,0.20}
\definecolor{gainlow}{rgb}{0.96,0.72,0.42}
\definecolor{gainghost}{rgb}{0.98,0.88,0.72}
\definecolor{overhigh}{rgb}{0.75,0.25,0.08}
\definecolor{overmed}{rgb}{0.88,0.45,0.15}
\definecolor{overlow}{rgb}{0.94,0.65,0.35}
\definecolor{overghost}{rgb}{0.97,0.84,0.68}

\begin{table}[htbp]
\centering
\caption{Per-task performance on 17 MLE-Bench Lite tasks with valid initial scores.}
\label{tab:perf_details}
\resizebox{\textwidth}{!}{
\begin{tabular}{ll rrr rr}
\toprule
\textbf{Competition} & \textbf{Dir.} & \textbf{Initial} & \textbf{Best} & \textbf{Rel. Impr.} & \textbf{Gain\%} & \textbf{Overcome} \\
\midrule
mlsp-2013-birds         & Higher & 0.86753 & 0.93870 & +0.07117 & \cellcolor{gainhigh}\textcolor{white}{102.33\%} & \cellcolor{overhigh}\textcolor{white}{85.71\%} \\
aptos2019-blindness     & Higher & 0.92106 & 0.96239 & +0.04133 & \cellcolor{gainhigh}\textcolor{white}{99.36\%}  & \cellcolor{overghost}20.00\% \\
nomad2018               & Lower  & 0.06033 & 0.05207 & +0.00826 & \cellcolor{gainhigh}\textcolor{white}{59.04\%}  & \cellcolor{overhigh}\textcolor{white}{88.89\%} \\
text-norm-english       & Higher & 0.99285 & 0.99575 & +0.00290 & \cellcolor{gainmed}42.21\%  & \cellcolor{overhigh}\textcolor{white}{86.70\%} \\
plant-pathology         & Higher & 0.98142 & 0.99350 & +0.01208 & \cellcolor{gainmed}40.48\%  & \cellcolor{overhigh}\textcolor{white}{80.00\%} \\
leaf-classification     & Lower  & 0.02367 & 0.00004 & +0.02363 & \cellcolor{gainmed}21.81\%  & \cellcolor{overhigh}\textcolor{white}{92.90\%} \\
spooky-author           & Lower  & 0.23456 & 0.19931 & +0.03525 & \cellcolor{gainlow}13.89\%  & \cellcolor{overmed}60.00\% \\
text-norm-russian       & Higher & 0.97232 & 0.97393 & +0.00161 & \cellcolor{gainlow}11.33\%  & \cellcolor{overmed}60.00\% \\
dogs-vs-cats            & Lower  & 0.01028 & 0.00331 & +0.00697 & \cellcolor{gainlow}8.36\%   & \cellcolor{overhigh}\textcolor{white}{87.50\%} \\
random-acts-of-pizza    & Higher & 0.62265 & 0.64955 & +0.02690 & \cellcolor{gainlow}7.09\%   & \cellcolor{overmed}76.92\% \\
dog-breed               & Lower  & 0.46888 & 0.44098 & +0.02790 & \cellcolor{gainghost}5.92\%  & \cellcolor{overlow}57.89\% \\
denoising-documents     & Lower  & 0.03622 & 0.03341 & +0.00281 & \cellcolor{gainghost}5.08\%  & \cellcolor{overghost}14.30\% \\
tps-may-2022            & Higher & 0.99477 & 0.99583 & +0.00106 & \cellcolor{gainghost}4.15\%  & \cellcolor{overhigh}\textcolor{white}{81.25\%} \\
whale-redux             & Higher & 0.98777 & 0.99109 & +0.00332 & \cellcolor{gainghost}2.67\%  & \cellcolor{overhigh}\textcolor{white}{88.00\%} \\
tps-dec-2021            & Higher & 0.96330 & 0.96338 & +0.00008 & \cellcolor{gainghost}2.52\%  & \cellcolor{overlow}58.80\% \\
aerial-cactus           & Higher & 0.99999 & 1.00000 & +0.00001 & \cellcolor{gainghost}0.00\%  & \cellcolor{overghost}40.00\% \\
insults-detect          & Higher & 0.95637 & 0.95637 & +0.00000 & \cellcolor{gainghost}0.00\%  & \cellcolor{overghost}0.00\% \\
\bottomrule
\end{tabular}
}
\end{table}

\subsection{Cost Analysis}
To assess the practical cost of DataMaster, we measure wall-clock time and API token consumption (input and output separately) across all MLE-Bench Lite runs. Table~\ref{tab:efficiency} compares DataMaster against baselines in terms of compute cost and resulting performance.

\begin{table}[t]
\centering
\caption{%
  Efficiency comparison on MLE-Bench Lite.
  Tokens are reported as average per-task input/output tokens;
  time is average wall-clock minutes per task.
  Performance columns reuse the main-table results.
}
\label{tab:efficiency}
\renewcommand{\arraystretch}{1.15}
\begin{tabular}{l rrrr cc}
\toprule
\textbf{Method} & \makecell{Input\\Tokens (K)} & \makecell{Output\\Tokens (K)} & \makecell{API Cost\\(\$)} & \makecell{Time\\(min)} & \makecell{Medal\\Rate} & \makecell{Gold\\Rate} \\
\midrule
Claude Code      & 522.5 & 13.3 & 0.09 & 127.6 & 36.36\% & 22.12\% \\
Codex            & 97.3  & 19.0 & 0.05 & 95.4  & 22.73\% & 18.18\% \\
ML-Master 2.0    & 349.0 & 113.9 & 0.29 & 669.6 & 40.91\% & 27.27\% \\
DatasetResearch  & 38.5  & 64.6  & 0.15 & 71.7  & 59.09\% & 27.27\% \\
\midrule
\textbf{DataMaster} & 2690.3 & 596.3 & 1.61 & 718.7 & 68.18\% & 45.45\% \\
\bottomrule
\end{tabular}
\end{table}

\begin{table}[htbp]
\centering
\caption{Task-wise MLE-Lite test scores and overcome rates. Direction indicates whether higher or lower scores are better for each competition metric.}
\label{tab:mle_lite_task_scores}
\resizebox{\linewidth}{!}{
\begin{tabular}{llrrr}
\toprule
\textbf{Competition} & \textbf{Direction} & \textbf{Test Initial Score} & \textbf{Test Best Score} & \textbf{Overcome Rate} \\
\midrule
aerial-cactus-identification & Higher & 0.99999 & 1.00000 & 40\% \\
aptos2019-blindness-detection & Higher & 0.92106 & 0.96239 & 20\% \\
mlsp-2013-birds & Higher & 0.86753 & 0.93870 & 85.71\% \\
new-york-city-taxi-fare-prediction & Lower & \texttt{None} & 4.73358 & 100\% \\
plant-pathology-2020-fgvc7 & Higher & 0.98142 & 0.99350 & 80\% \\
ranzcr-clip-catheter-line-classification & Higher & \texttt{None} & 0.89733 & \texttt{None} \\
spooky-author-identification & Lower & 0.23456 & 0.19931 & 60\% \\
detecting-insults-in-social-commentary & Higher & 0.95637 & 0.95637 & 0.00\% \\
dog-breed-identification & Lower & 0.46888 & 0.44098 & 57.89\% \\
dogs-vs-cats-redux-kernels-edition & Lower & 0.01028 & 0.00331 & 87.50\% \\
histopathologic-cancer-detection & Higher & \texttt{None} & 0.99788 & 100\% \\
jigsaw-toxic-comment-classification-challenge & Higher & \texttt{None} & 0.98663 & \texttt{None} \\
nomad2018-predict-transparent-conductors & Lower & 0.06033 & 0.05207 & 88.89\% \\
random-acts-of-pizza & Higher & 0.62265 & 0.64955 & 76.92\% \\
text-normalization-challenge-english-language & Higher & 0.99285 & 0.99575 & 86.70\% \\
denoising-dirty-documents & Lower & 0.03622 & 0.03341 & 14.30\% \\
leaf-classification & Lower & 0.02367 & 0.00004 & 92.90\% \\
siim-isic-melanoma-classification & Higher & 0.90303 & 0.91516 & 92.86\% \\
tabular-playground-series-dec-2021 & Higher & 0.96330 & 0.96338 & 58.80\% \\
tabular-playground-series-may-2022 & Higher & 0.99477 & 0.99583 & 81.25\% \\
text-normalization-challenge-russian-language & Higher & 0.97232 & 0.97393 & 60\% \\
the-icml-2013-whale-challenge-right-whale-redux & Higher & 0.98777 & 0.99109 & 88\% \\
\bottomrule
\end{tabular}
}
\end{table}

\begin{table}[htbp]
\centering
\caption{Average MLE-Lite search-tree and trajectory-step statistics by competition category.}
\label{tab:mle_lite_category_nonscore}
\resizebox{\linewidth}{!}{
\begin{tabular}{lrrrrrrr}
\toprule
\textbf{Metric}
& \textbf{Image Cls.}
& \textbf{Tabular}
& \textbf{Text Cls.}
& \textbf{Audio Cls.}
& \textbf{Image Reg.}
& \textbf{Seq-to-Seq}
& \textbf{Image-to-Image} \\
\midrule
\texttt{node\_count} & 29.6250 & 34.5000 & 89.5000 & 83.5000 & 22.0000 & 33.0000 & 16.0000 \\
\texttt{red\_node\_count} & 16.0000 & 17.7500 & 66.5000 & 61.5000 & 14.0000 & 15.5000 & 8.0000 \\
\texttt{black\_node\_count} & 12.6250 & 15.7500 & 22.0000 & 21.0000 & 7.0000 & 16.5000 & 7.0000 \\
\texttt{initial\_node\_count} & 1.0000 & 1.0000 & 1.0000 & 1.0000 & 1.0000 & 1.0000 & 1.0000 \\
\texttt{avg\_steps\_per\_node} & 30.7981 & 35.1650 & 22.5211 & 20.9376 & 36.6818 & 39.0038 & 40.7500 \\
\texttt{red\_avg\_steps\_per\_node} & 28.7049 & 31.9904 & 22.1431 & 18.0587 & 37.8571 & 39.8364 & 39.3750 \\
\texttt{black\_avg\_steps\_per\_node} & 35.5963 & 40.5089 & 25.3585 & 28.5417 & 36.7143 & 39.5919 & 46.2857 \\
\bottomrule
\end{tabular}
}
\end{table}
\clearpage

\section{Scheduling Policy}
\label{app:scheduling_details}

This appendix provides the full specification of the UCB1-based scheduling policy introduced in Section~\ref{subsec:growth_scheduling}.

\subsection{Reward Design and Backpropagation}

The DataTree contains two types of executable nodes with fundamentally different outputs: red nodes produce candidate datasets (no training score), while black nodes produce downstream evaluation metrics. To unify them under a single scheduling framework, we define the immediate reward as
\[
r_v =
\begin{cases}
\epsilon, & \text{if } v \text{ is a red node and completes without error},\\
y_v,      & \text{if } v \text{ is a black node with evaluation score } y_v,\\
0,        & \text{otherwise (buggy or failed nodes)},
\end{cases}
\]
where $\epsilon$ is a small positive constant (we use $\epsilon = 0.01$ in all experiments). This ensures that successful data-discovery steps receive non-zero credit without dominating the exploitation signal from actual training outcomes.

Upon completion of node $v$ with reward $r_v$, the statistics are updated along the entire root path. Let $\mathrm{Path}(v) = (v, p(v), p(p(v)), \ldots, \mathrm{root})$ denote the sequence of nodes from $v$ to the root. For every $u \in \mathrm{Path}(v)$:
\begin{align}
N_u &\leftarrow N_u + 1, \label{eq:visit_update}\\
R_u &\leftarrow R_u + r_v. \label{eq:reward_update}
\end{align}
That is, both $v$ itself and all of its ancestors receive the same reward increment and visit increment. This backpropagation mechanism allows parent and grandparent nodes to accumulate evidence about the quality of their subtrees, enabling the scheduler to compare branches at different depths on a common scale.

\subsection{Node Initialization}

When a new frontier node $v$ is created by the growth policy, its statistics are initialized as
\[
N_v = 0, \quad R_v = 0.
\]
Since $N_v = 0$ would make the UCB1 score in Eq.~\eqref{eq:score} undefined (division by zero), we handle unvisited nodes as a special case:
\[
\mathrm{Score}(v) =
\begin{cases}
+\infty, & \text{if } N_v = 0, \\[4pt]
\dfrac{R_v}{N_v} + c_t\sqrt{\dfrac{\log N_{p(v)}}{N_v}}, & \text{if } N_v \geq 1.
\end{cases}
\]
This optimistic initialization guarantees that every newly generated frontier node is selected for execution before any previously visited node is revisited, ensuring that no branch is left completely unexplored. Once a node has been visited at least once ($N_v \geq 1$), both the exploitation term $R_v / N_v$ and the exploration term are well-defined, and the standard UCB1 formula applies.

\subsection{Exploration Coefficient Decay}
\label{app:decay}

The exploration coefficient $c_t$ controls the trade-off between exploiting high-reward branches and exploring under-visited ones. Let $t$ denote the number of node executions completed so far (i.e., the global step counter). We adopt a piecewise linear decay schedule that transitions through three phases over the total budget of $T$ rounds:
\[
c_t =
\begin{cases}
c_0,                                         & t < t_1, \\[4pt]
\max\!\bigl(c_0 - \alpha\,(t - t_1),\; c_{\min}\bigr), & t_1 \le t \le t_2, \\[4pt]
c_{\min},                                    & t > t_2,
\end{cases}
\]
where:
\begin{itemize}
    \item $c_0$ is the initial exploration constant,
    \item $\alpha$ is the linear decay rate per step during the transition phase,
    \item $c_{\min}$ is the lower bound preventing purely greedy behavior,
    \item $t_1 = \lfloor p_1 \cdot T \rfloor$ and $t_2 = \lfloor p_2 \cdot T \rfloor$ are the phase boundaries, defined as fractions of the total budget.
\end{itemize}

The three phases serve distinct purposes:
\begin{enumerate}
    \item \textbf{Exploration phase} ($t < t_1$): $c_t$ remains at its maximum value $c_0$, encouraging broad coverage of the frontier when little information is available.
    \item \textbf{Transition phase} ($t_1 \leq t \leq t_2$): $c_t$ decays linearly, gradually shifting priority toward branches that have already demonstrated strong performance.
    \item \textbf{Exploitation phase} ($t > t_2$): $c_t$ is clamped at $c_{\min}$, concentrating the remaining budget on the most promising branches while retaining a minimal exploration incentive to avoid premature convergence.
\end{enumerate}

\subsection{Hyperparameter Settings}
\label{app:hyperparams}

Table~\ref{tab:scheduling_hyperparams} lists the default hyperparameters used in all experiments.

\begin{table}[h]
\centering
\caption{Default scheduling hyperparameters.}
\label{tab:scheduling_hyperparams}
\begin{tabular}{lll}
\toprule
\textbf{Symbol} & \textbf{Description} & \textbf{Value} \\
\midrule
$c_0$ & Initial exploration constant & $1.414\;(\approx\!\sqrt{2})$ \\
$c_{\min}$ & Lower bound on $c_t$ & $0.5$ \\
$\alpha$ & Linear decay rate & $0.01$ \\
$p_1$ & Exploration phase ratio & $0.3$ \\
$p_2$ & Transition end ratio & $0.7$ \\
$\epsilon$ & Red-node reward & $0.01$ \\
$T$ & Maximum rounds (budget) & $40$ \\
\midrule
\multicolumn{3}{l}{\textit{Tree structure parameters}} \\
\midrule
\texttt{num\_red} & Red nodes per expansion & $1$ \\
\texttt{num\_black} & Initial black nodes per red & $5$ \\
\texttt{max\_black\_per\_red} & Maximum black nodes per red & $5$ \\
\bottomrule
\end{tabular}
\end{table}

The choice of $c_0 = \sqrt{2}$ follows the theoretical recommendation for UCB1 when rewards lie in $[0,1]$. The phase ratios $(p_1, p_2) = (0.3, 0.7)$ allocate roughly the first $30\%$ of the budget to pure exploration, the middle $40\%$ to transition, and the final $30\%$ to exploitation. The lower bound $c_{\min} = 0.5$ ensures that even in the late exploitation phase, a branch with significantly fewer visits than its siblings retains a meaningful chance of being selected.

\subsection{Alternative Decay Schedules}

While we use the piecewise schedule by default, our implementation also supports two alternative decay strategies for ablation purposes:

\paragraph{Linear decay.}
\[
c_t = \max\!\bigl(c_0 - \alpha \cdot t,\; c_{\min}\bigr).
\]
This provides a steady, monotonic decrease from the first step onward, without a dedicated exploration phase.

\paragraph{Exponential decay.}
\[
c_t = \max\!\bigl(c_0 \cdot \gamma^t,\; c_{\min}\bigr),
\]
where $\gamma \in (0,1)$ is the decay factor (default $\gamma = 0.99$). Exponential decay reduces exploration aggressively in early steps but converges more slowly to $c_{\min}$ compared to linear decay.

\subsection{Priority Queue Implementation}

The scheduler maintains a max-priority queue over the frontier $\mathcal{N}$. Each time a node completes and new children are generated by the growth policy, the new nodes are inserted with $\mathrm{Score}(v) = +\infty$. After node completion triggers backpropagation (Eqs.~\ref{eq:visit_update}--\ref{eq:reward_update}), existing frontier nodes are not re-scored eagerly; instead, scores are computed lazily at selection time. The next node $v_{\mathrm{next}}$ is popped from the queue, and execution proceeds. This design keeps scheduling overhead constant per step regardless of frontier size.

\section{Case Study}

\subsection{Branch Concentration Bias in Leaf Classification}

We further analyze tree-growth asymmetry using the \texttt{leaf-classification} task as a case study. We reconstruct the search tree from the UCT node logs using each node's identifier and parent identifier. After excluding the artificial root and initial node, we treat the children of the initial node as root-level branches.

For each search step \(t\), let \(n_i(t)\) be the number of expanded nodes assigned to branch \(i\), and let \(K\) be the total number of root-level branches. We first convert branch sizes into proportions:
\[
p_i(t) = \frac{n_i(t)}{\sum_{j=1}^{K} n_j(t)}.
\]
We then measure branch concentration using the normalized Herfindahl-style index:
\[
\mathrm{Bias}(t)
=
\frac{
\sum_{i=1}^{K} p_i(t)^2 - \frac{1}{K}
}{
1 - \frac{1}{K}
}.
\]
This value ranges from 0 to 1, where 0 indicates uniform expansion across branches and 1 indicates that all expanded nodes are concentrated in a single branch.

\begin{figure}[htbp]
    \centering
    \includegraphics[width=0.75\linewidth]{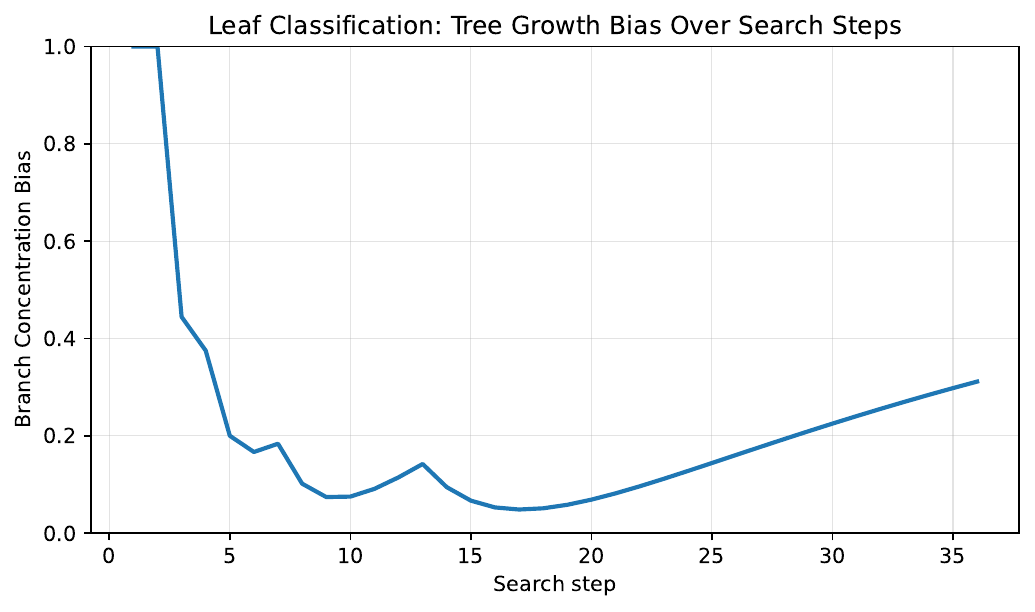}
    \caption{Branch Concentration Bias over search steps for the \texttt{leaf-classification} task. Higher values indicate more asymmetric tree growth. The curve initially decreases as the search expands across multiple branches, but later rises as one branch receives a larger share of subsequent expansions.}
    \label{fig:leaf_branch_bias}
\end{figure}

The \texttt{leaf-classification} tree contains 38 extracted nodes and 5 root-level branches after removing the artificial root and initial node. The final branch sizes are 23, 6, 3, 2, and 2 nodes, showing a clear imbalance in how expansion is allocated. As shown in Fig.~\ref{fig:leaf_branch_bias}, the bias decreases during the early search steps, suggesting initial diversification across branches. However, the later increase indicates that the agent increasingly concentrates expansion on a dominant branch. This supports the observation that the search process is not uniform; instead, it exhibits branch-level concentration as the tree develops.

\subsection{Red--Black Exploration Ratio Analysis}

To further examine the structural behavior of the search process, we compare Red-node and Black-node executions across competition categories. For each task, we compute two ratios. The first is the executed-node ratio:
\[
R_{\mathrm{node}}
=
\frac{N_{\mathrm{Red}}}{N_{\mathrm{Black}}},
\]
where \(N_{\mathrm{Red}}\) and \(N_{\mathrm{Black}}\) denote the numbers of executed Red and Black trajectory nodes. The second is the tool-intensity ratio:
\[
R_{\mathrm{tool}}
=
\frac{T_{\mathrm{Red}} / N_{\mathrm{Red}}}
     {T_{\mathrm{Black}} / N_{\mathrm{Black}}},
\]
where \(T_{\mathrm{Red}}\) and \(T_{\mathrm{Black}}\) denote the total numbers of tool calls made by executed Red and Black nodes. Ratios are first computed at the task level and then averaged within each competition category.

\begin{figure}[htbp]
    \centering
    \includegraphics[width=0.95\linewidth]{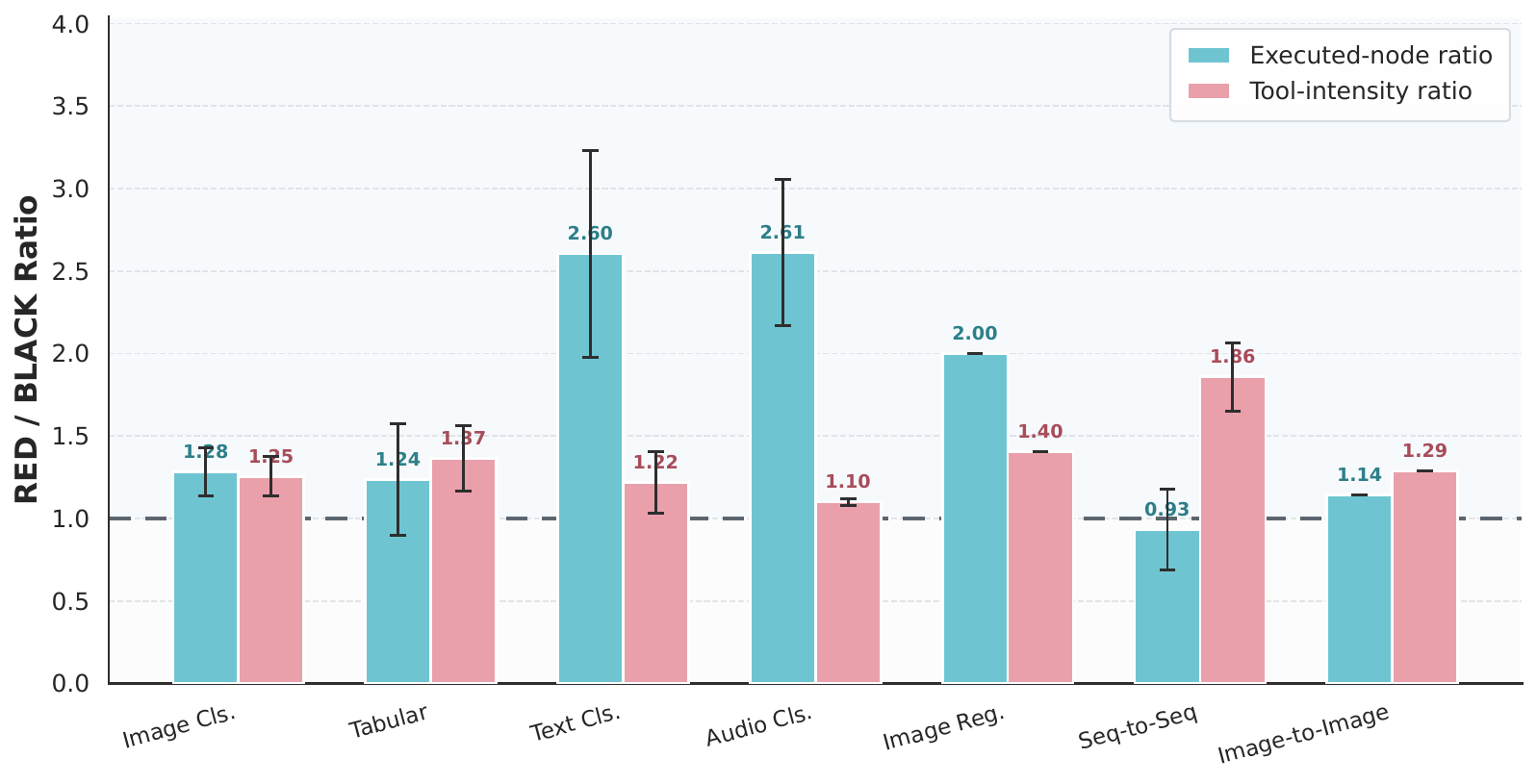}
    \caption{Red--Black exploration ratios by competition category. The blue bars show the executed-node ratio, while the pink bars show the tool-intensity ratio. The dashed horizontal line at \(y=1\) indicates parity between Red and Black nodes. Values above 1 indicate Red-node dominance, whereas values below 1 indicate Black-node dominance. Error bars denote the standard error of the mean across tasks within each category.}
    \label{fig:red_black_stage_ratio}
\end{figure}

Overall, most competition categories show executed-node ratios above 1, suggesting that the agent tends to allocate more executed nodes to Red-node exploration than to Black-node implementation. The tool-intensity ratio is also generally near or above 1, but it is usually less pronounced than the executed-node ratio. This indicates that the Red--Black imbalance is primarily structural, reflected in the number of executed nodes, while per-node tool-use intensity provides a secondary and more task-dependent source of imbalance.

\subsection{Text Embedding Distribution for Spooky Author Identification}

To examine the semantic alignment between the available training/validation data and the public test set, we visualize the \textit{spooky-author-identification} task using a t-SNE projection. We sample 1,000 examples from the training/validation data and 1,000 examples from the public test set, using only the text field. Each text instance is embedded with Qwen3-Embedding-0.6B, L2-normalized, and projected into two dimensions using t-SNE with perplexity 30 and a fixed random seed.

\begin{figure}[H]
    \centering
    \includegraphics[width=0.72\linewidth]{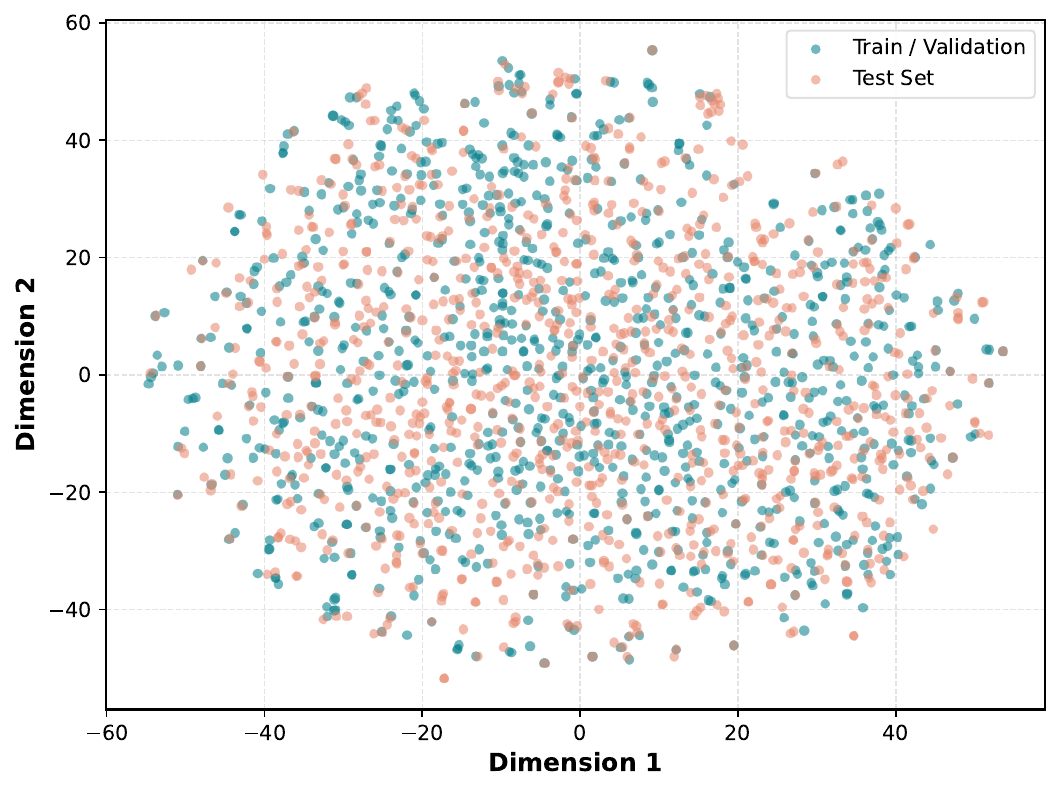}
    \caption{Text embedding t-SNE visualization for the \textit{spooky-author-identification} task. Each point corresponds to one text example, with colors indicating whether the example comes from the training/validation data or the public test set. The substantial overlap between the two groups suggests that the public test set is broadly aligned with the available training/validation distribution in the embedding space. This alignment also helps explain why task-relevant public text resources can potentially support downstream modeling and evaluation.}
    \label{fig:spooky-tsne-embeddings}
\end{figure}

The projection shows that training/validation and test examples are highly intermixed, with no clear separation between the two groups. This pattern suggests that the public test distribution is close to the available training/validation distribution under the chosen embedding model. More broadly, it indicates that publicly available text data may contain useful signals for downstream tasks, which helps motivate retrieval- and search-based strategies in agentic model development.

\subsection{Initial Code Examples}
\label{sec:init_code_examples}

We present the initial code used in the dog-breed-identification task as a representative example of the two code settings in Section~\ref{sec:ablation}.
Listing~\ref{lst:init_code} shows the \emph{DataLoader} component, which handles data loading, label encoding, train/validation splitting, and data augmentation and \emph{Algorithm} component, which implements model construction, the training loop and validation.
In the \textbf{Full-code} setting, both components are provided; in the \textbf{Algo-only} setting, only the Algorithm component is included and the agent must construct the DataLoader from scratch.

\begin{lstlisting}[style=python, xleftmargin=2em, caption={Initial code structure for dog-breed-identification (both DataLoader and Algorithm components).}, label={lst:init_code}]
# ==================== DataLoader Component ====================
class DogDataset(Dataset):
    def __init__(self, ids, labels, img_dir, transform, is_train):
        ...
    def __getitem__(self, idx):
        image = Image.open(...).convert("RGB")
        if self.transform: image = self.transform(image)
        return (image, self.labels[idx]) if self.is_train else image

class MyDataLoader(BaseDataLoader):
    def __init__(self, batch_size=32, num_workers=8,
                 input_dir="./input/", **kwargs): ...

    def setup(self):
        # Load and encode labels
        labels_df = pd.read_csv(os.path.join(self.input_dir, "labels.csv"))
        le = LabelEncoder()
        labels_df["breed_encoded"] = le.fit_transform(labels_df["breed"])
        # Train/val split (fixed val.csv or stratified 80/20)
        ...
        # Augmentation: Resize(384), RandomCrop(320), Flip, ColorJitter
        self.train_transform = transforms.Compose([...])
        self.val_transform = transforms.Compose([...])
        # Create DogDataset + DataLoader for train/val/test
        ...

# ==================== Algorithm Component ====================
class MyAlgorithm(BaseAlgorithm):
    def __init__(self, model_name='efficientnet_b3',
                 lr=1e-4, epochs=10, **kwargs): ...

    def fit(self, train_data):
        # Build pretrained EfficientNet-B3 via timm
        self.model = timm.create_model(self.model_name,
            pretrained=True, num_classes=num_classes)
        # AdamW optimizer + CosineAnnealingLR scheduler
        ...
        # Training loop with validation; save best checkpoint
        for epoch in range(self.epochs):
            # forward -> loss -> backward -> step
            ...
            # validate and save best model
            ...

    def predict(self, test_data):
        self.model.eval()
        with torch.no_grad():
            for images in test_loader:
                # TTA: average(original, horizontal flip)
                outputs = self.model(images)
                outputs_flip = self.model(torch.flip(images, [3]))
                probs = (softmax(outputs) + softmax(outputs_flip)) / 2
                ...
        # Create submission DataFrame
        return pd.DataFrame({'id': test_ids, 'breed': predicted})
\end{lstlisting}


\end{document}